\documentclass[9pt,technote]{IEEEtran}
\ifCLASSINFOpdf
\else
\fi

\usepackage{caption}
\ifCLASSOPTIONcompsoc
 \usepackage[caption=false,font=normalsize,labelfont=sf,textfont=sf]{subfig}
\else
 \usepackage[caption=false,font=footnotesize]{subfig}
\fi

\usepackage{times}

\usepackage{multicol}
\usepackage[bookmarks=true]{hyperref}

\usepackage{graphicx,import}
\usepackage{mathtools, amssymb, amsthm}
\usepackage{paralist}
\usepackage[pdf]{svg}
\usepackage{cleveref}
\usepackage{commath}

\newtheorem{assumption}{\bf{Assumption}}
\newtheorem{definition}{\bf{Definition}}
\newtheorem{boldLemma}{\bf{Lemma}}

\newtheorem{Remark}{Remark}

\newtheorem{hypothesis}{Hypothesis}

\makeatletter
\newsavebox\myboxA
\newsavebox\myboxB
\newlength\mylenA
\newcommand*\xoverline[2][0.75]{%
    \sbox{\myboxA}{$\m@th#2$}%
    \setbox\myboxB\null
    \ht\myboxB=\ht\myboxA%
    \dp\myboxB=\dp\myboxA%
    \wd\myboxB=#1\wd\myboxA
    \sbox\myboxB{$\m@th\overline{\copy\myboxB}$}
    \setlength\mylenA{\the\wd\myboxA}
    \addtolength\mylenA{-\the\wd\myboxB}%
    \ifdim\wd\myboxB<\wd\myboxA%
       \rlap{\hskip 0.5\mylenA\usebox\myboxB}{\usebox\myboxA}%
    \else
        \hskip -0.5\mylenA\rlap{\usebox\myboxA}{\hskip 0.5\mylenA\usebox\myboxB}%
    \fi}
\makeatother

\makeatletter
\newcommand{\subalign}[1]{%
  \vcenter{%
    \Let@ \restore@math@cr \default@tag
    \baselineskip\fontdimen10 \scriptfont\tw@
    \advance\baselineskip\fontdimen12 \scriptfont\tw@
    \lineskip\thr@@\fontdimen8 \scriptfont\thr@@
    \lineskiplimit\lineskip
    \ialign{\hfil$\m@th\scriptstyle##$&$\m@th\scriptstyle{}##$\crcr
      #1\crcr
    }%
  }
}
\makeatother

\graphicspath{{imgs/}}

\begin{document}
%
\title{
The Static Center of Pressure Sensitivity:\\
a further Criterion to assess Contact Stability\\ and Balancing Controllers
}

%
%
%

%
%

\author{Francesco~Romano, Daniele~Pucci,  Silvio~Traversaro, Francesco~Nori
\thanks{*This work was supported by the FP7 EU project CoDyCo (No. 600716 ICT 2011.2.1 Cognitive Systems and Robotics) and Koroibot (No. 611909 ICT-2013.2.1 Cognitive Systems and Robotics).}
\thanks{The authors are with the iCub Facility department, Istituto Italiano di Tecnologia,
        Via Morego 30, Genoa, Italy.
        {\tt\small name.surname@iit.it}}
\thanks{Manuscript submitted October 3, 2016.}%
}

\markboth{Journal of \LaTeX\ Class Files,~Vol.~14, No.~8, August~2015}%
{Shell \MakeLowercase{\textit{et al.}}: Bare Demo of IEEEtran.cls for IEEE Journals}
%



\maketitle

\begin{abstract}
Legged locomotion has received increasing attention from the robotics community.
In this respect, contact stability plays a critical role in ensuring that robots maintain balance, and it is a key element for balancing and walking controllers.
The Center of Pressure is a contact stability criterion that defines a point that must be kept strictly inside the support polygon in order to ensure \emph{postural stability}.
In this paper, we introduce the concept of the sensitivity of the \emph{static} center of pressure: roughly speaking, the rate of change of the center of pressure with respect to the system equilibrium configurations. 
This new concept can be used as an additional criterion to assess the \emph{robustness} of the contact stability.
We show how the sensitivity of the center of pressure can also be used as a metric to assess balancing controllers 
by considering two state-of-the-art control strategies.
The analytical analysis is performed on a simplified model, and validated during balancing tasks on the iCub humanoid robot.
\end{abstract}


%
\IEEEpeerreviewmaketitle

\section{Introduction}

\IEEEPARstart{I}{n the last} decades, legged robots have become increasingly more important.
Climbing stairs, avoiding obstacles, and exploiting multiple contacts are only a few advantages that humanoid robots have over wheeled systems.
These advantages have been the main motivations behind the attention they have received from the robotics community.
Legged locomotion, on the other hand, arises interesting issues, which must be solved before humanoids can be successfully deployed in the human environment.
The recent DARPA Robotics Challenge has clearly highlighted these difficulties \cite{DRC-what-happened}:
most of the failures during trials were due to a non perfect balancing or walking.
This paper introduces a new metric, called \emph{sensitivity} of the static center of pressure, which can be further used to analyse contact stability.

%


Nowadays there exist two main methodologies to implement balancing and walking controllers on bipedal robots: 
the first class is based robot simplified models \cite{Kajita2001,Lee07reactionmass,Pratt2006LIP} complemented with the use of inverse kinematics to bridge the gap between the simplified model and the real one \cite{Kajita2003,Stephens2006,Pratt01092012}.
The second class is based on the full dynamic model of the robot \cite{Herzog2014,Sentis2010,Nori2015,Righetti01032013}
and it has been successfully exploited for the synthesis of instantaneous controllers.
In this regard,
a special attention is paid to controlling the interaction forces between the robot and the environment \cite{Herzog2014,Ott2011human}.

Given the complex nature of 
humanoid robots, researchers have progressively shifted from using classic nonlinear control techniques to adopting optimisation-based solutions \cite{Ott2011,Koolen2016,Wensing2013,Hopkins2015b}.
These optimisation-based control techniques allow one to specify, at high level, the tasks to be accomplished by the robot: the underlining solver is then in charge of finding a feasible solution to the optimisation problem.
One disadvantage of this process is that properties of the solution can be more difficult to find. If some properties have to be enforced, they can be specified by means of constraints or optimisation criteria.
But how this high-level criteria influence the solution is not trivial and must be analysed on a case-by-case basis.
In this paper, we analyse two state-of-the-art balancing controllers commonly found in literature (\cite{Herzog2014, Hopkins2015b,Lee2010,Liu2015,Nori2015} and \cite{Nava16,PucciVideo2016}), and we analyse them by means of the center of pressure sensitivity here introduced: we
show how these criteria yield different performance when the robot balances on its two feet.

Maintaining the support foot (or feet) \emph{stable} is of paramount importance for balancing and walking.
As a result, various \emph{postural stability} criteria  analysing the stance foot dynamics have been proposed in the robotics and human locomotion literature.
These criteria
focus on the force and torques exchanged at the ground-foot interface.

The Ground projection of the Center of Mass (GCoM) can be applied if the robot dynamics is at equilibrium:
it states that to assess the \emph{postural stability} of the biped, the projection on the ground of the CoM must lie inside the convex hull of the support feet.

The Center of Pressure (CoP), also known as Zero Moment Point (ZMP) \cite{vukobratovic2004zero,Goswami01061999}, can be applied as soon as the static assumption is relaxed.
The center of pressure can be viewed as the point on the ground surface where the resultant of the normal contact force acts. 
Because of the unilaterality of the normal component of the contact forces, the center of pressure must lie inside the support polygon.
It is worth noting that the CoP coincides with the GCoM when the stationarity hypothesis is satisfied \cite{Goswami01061999}.

Finally, the Foot Rotation Indicator (FRI) point has been recently introduced \cite{Goswami01061999}: 
when the foot does not rotate, the FRI point coincides with the CoP, and lies inside the support polygon.
Differently than the CoP, the FRI can exit the support polygon, which occurs when the foot starts rotating.

All the previously described postural stability criteria are not directly linked to a specific robot dynamical model or balancing controller implementation, thus being useful for a multitude of control solutions.
On the other hand, one of  these criteria (e.g. CoP, FRI, etc) must be taken into account during the synthesis of postural controllers to ensure stable foot planting and postural stability of the whole robot.

In this work, we introduce the concept of the sensitivity of the static center of pressure: roughly speaking the rate of change of the center of pressure versus the system equilibrium configuration.
We show how this new concept can further characterise the stability of a contact, thus complementing the notion of center of pressure.

We also show that the sensitivity of the center of pressure can be used to evaluate the performances of different balancing controllers.
In this regard we analyse two state-of-the-art controllers:
they both stabilise the robot linear and angular momentum as control objective, assuming the contact forces and torques as virtual control inputs.
Because the system is overdetermined in presence of more than one contact, the resulting force redundancy can be exploited in different ways.
In particular, the two state-of-the-art controllers differ in the criterion used to resolve the  redundancy:
the minimisation of the contact wrenches norm \cite{Herzog2014, Hopkins2015b,Lee2010,Liu2015,Nori2015,Hyon2007,WENSING2013b} in one case and the minimisation of the internal joint torques norm \cite{Nava16,PucciVideo2016} in the other:
we show that the two criteria exhibit different static center of pressure sensitivities.
We can thus exploit the introduced metric to discriminate between two, apparently similar, controllers.


To simplify the analytical analysis, we model the lower body of a humanoid robot while balancing as a planar four-bar linkage mechanism and 
we further extend and apply the analysis on the iCub humanoid robot so as to assess the validity of the simplified model.

The paper is structured as follows. 
Section~\ref{sec:background} introduces the dynamical model of a free floating mechanical system and the notation used throughout the paper and Section~\ref{sec:scop} describes the concept of \emph{sensitivity of static center of pressure}.
The sensitivity of the static center of pressure is applied first to a four-bar linkage model in Section~\ref{sec:four-bar} and then to the iCub humanoid robot in Section~\ref{sec:experiments}.
Finally Section~\ref{sec:conclusions} draws the conclusions.

\section{Background}
\label{sec:background}

\subsection{Notation}
Throughout the paper we will use the following definitions.
\begin{itemize}
    \item $\mathcal{I}$ denotes an inertial frame, with its $z$ axis pointing against the gravity. We denote with $g$ the gravitational constant.
    \item $e_i \in \mathbb{R}^m$ is the canonical vector, consisting of all zeros but the $i$-th component, which is one.
    \item Given two orientation frames $A$ and $B$, and vectors of coordinates expressed in these frames, i.e. $\prescript{A}{}p$ and $\prescript{B}{}p$, respectively, the rotation matrix 
    $\prescript{A}{}R_B$ is such that $\prescript{A}{}p = \prescript{A}{}R_B  \prescript{B}{}p$. 
    \item $1_n \in \mathbb{R}^{n \times n}$ is the identity matrix of size $n$; $0_{m \times n} \in \mathbb{R}^{m \times n}$ is the zero matrix of size $m \times n$ and $0_{n } = 0_{n \times 1}$.
    \item We denote with $S(x) \in \mathbb{R}^{3 \times 3}$ the skew-symmetric matrix such that $S(x)y = x \times y$, where $\times$ denotes the cross product operator in $\mathbb{R}^3$. If $x=({\tilde{x}}^\top,0)^\top$ and $y=({\tilde{y}}^\top,0)^\top$, with ${\tilde{x}},\tilde{y} \in \mathbb{R}^2$ one has $S(\tilde{x})\tilde{y} = -\tilde{x}^\top S \tilde{y} e_3$, with 
    $S = \begin{bmatrix}
        0 & -1 \\ 1 & 0
    \end{bmatrix}$.
\end{itemize}

\subsection{System modelling}
We assume that the robot is composed of $n+1$ rigid bodies -- called links -- connected by $n$ joints with one degree of freedom each. In addition, we also assume   that the multi-body system is \emph{free floating}, i.e. none of the links has an \emph{a priori} constant pose with respect to the inertial frame. This implies that  the multi-body system possesses $n~+~6$ degrees of freedom. The 
configuration space of the multi-body system can then be characterised by the \emph{position} and the \emph{orientation} of a frame attached to a robot's link -- called 
\emph{base frame} $\mathcal{B}$ -- and the joint configurations. More precisely, the robot configuration space  is defined by
\begin{equation*}
    \mathbb{Q} = \mathbb{R}^3 \times SO(3) \times \mathbb{R}^n.
\end{equation*}
An element of the set $\mathbb{Q}$ is then a triplet $q = (\prescript{\mathcal{I}}{}p_{\mathcal{B}},\prescript{\mathcal{I}}{}R_{\mathcal{B}},q_j)$, where $(\prescript{\mathcal{I}}{}p_{\mathcal{B}},\prescript{\mathcal{I}}{}R_{\mathcal{B}})$ denotes the origin  and orientation of the \emph{base frame} expressed in the inertial frame, and $q_j$ denotes the \emph{joint angles}. 
The \emph{velocity} of the multi-body system can then be characterised by the \emph{algebra} $\mathbb{V}$ of $\mathbb{Q}$ defined by:
    \[\mathbb{V} = \mathbb{R}^3 \times \mathbb{R}^3 \times \mathbb{R}^n.\]
An element of $\mathbb{V}$ is then a triplet $\nu = ( ^\mathcal{I}\dot{ p}_{\mathcal{B}},^\mathcal{I}\omega_{\mathcal{B}},\dot{q}_j)$, where $^\mathcal{I}\omega_{\mathcal{B}}$ is the angular velocity of the base frame expressed w.r.t. the inertial frame, i.e. $^\mathcal{I}\dot{R}_{\mathcal{B}} = S(^\mathcal{I}\omega_{\mathcal{B}})^\mathcal{I}{R}_{\mathcal{B}}$.


We also assume that the robot is interacting with the environment through $n_c$ distinct contacts. 
Applying  the Euler-Poincar\'e formalism to the multi-body system  yields the following equations of motion \cite[Ch. 13.5]{Marsden2010}\footnote{The Euler-Lagrange's formulation can be applied only to mechanical systems evolving in vector spaces. The Euler-Poincar\'e equations, instead, are valid for mechanical systems evolving in arbitrary Lie groups.}: 
\begin{align}
    \label{eq:system_dynamics}
       {M}(q)\dot{{\nu}} + {C}(q, {\nu}) {\nu} + {G}(q) =  B \tau + \sum_{k = 1}^{n_c} {J}^\top_{\mathcal{C}_k} f_k
\end{align}
where ${M} \in \mathbb{R}^{n+6 \times n+6}$ is the mass matrix, ${C} \in \mathbb{R}^{n+6 \times n+6}$ is the Coriolis matrix, ${G} \in \mathbb{R}^{n+6}$ is the gravity term, $B = (0_{n\times 6} , 1_n)^\top$ is a selector matrix, $\tau$ are the internal actuation torques, and $f_k$  denotes an external wrench applied by the environment to the link of the $k$-th contact. We assume that 
the external wrench is associated with a frame $\mathcal{C}_k$, which is attached to the robot's link where the wrench acts on.
Then,  the external wrench $f_k$ is expressed in a frame whose orientation coincides with that of the inertial frame $\mathcal{I}$, but whose origin is the  origin of $\mathcal{C}_k$.

The Jacobian ${J}_k= {J}_k(q)$ is the map between the robot's velocity ${\nu}$ and the linear and angular velocity $ ^\mathcal{I}v_{\mathcal{C}_k} := (^\mathcal{I}\dot{ p}_{\mathcal{C}_k},^\mathcal{I}\omega_{\mathcal{C}_k})$ of the frame $\mathcal{C}_k$, i.e.
\begin{align} 
^\mathcal{I}v_{\mathcal{C}_k} = {J}_{\mathcal{C}_k}(q) {\nu}.
\end{align}
The Jacobian has the following structure. 
\begin{IEEEeqnarray}{RCLRLL}
\label{eqn:jacobian}
{J}_{\mathcal{C}_k}(q) &=& \begin{bmatrix} {J}_{\mathcal{C}_k}^b(q) & {J}_{\mathcal{C}_k}^j(q)\end{bmatrix} &\in& \mathbb{R}^{6\times n+6}, \IEEEyessubnumber \\ 
 {J}_{\mathcal{C}_k}^b(q) &=& 
 \begin{bmatrix}
 1_3 & -S(\prescript{\mathcal{I}}{}p_{\mathcal{C}_k}-\prescript{\mathcal{I}}{}p_{\mathcal{B}})\\ 
 0_{3\times3} & 1_3 \\ 
 \end{bmatrix} &\in& \mathbb{R}^{6\times6} . \IEEEyessubnumber
\end{IEEEeqnarray}

Lastly, we assume that holonomic constraints may be enforced on the robot due to rigid contacts with the environment. 
These constraints are  modelled as a kinematic constraints that forbid any motion of the frame $\mathcal{C}_k$, i.e. ${J}_{\mathcal{C}_k}(q) {\nu} = 0$.
\section{Sensitivity of the Static Center-of-Pressure}
\label{sec:scop}
This section discusses a property of any center of pressure associated with 
  rigid, flat contacts and quasi-static robot's configurations. We shall see in Section~\ref{sec:four-bar} that this property --~called \emph{static center of pressure sensitivity}~-- can be used as an element of evaluation of contact stability and  balancing controllers for humanoid robots.
\subsection{The sensitivity of the static center of pressure in a case study: the four-bar linkage case }
  \begin{figure}[t]
      \centering{
          \def\svgwidth{\columnwidth}
          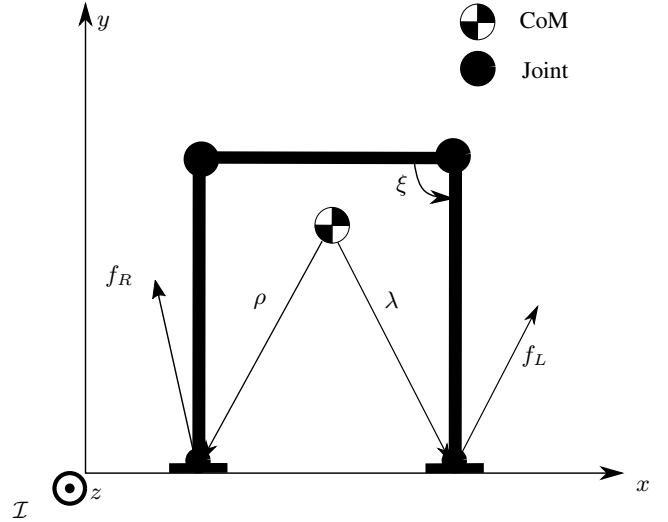
      \caption{Planar four-bar linkage exploiting two rigid contact to stand}
      \label{fig:fourBarLinkageStanding}
      }
  \end{figure}

To provide the reader with an introduction to the \emph{sensitivity} of the static center of pressure, focus on Figure~\ref{fig:fourBarLinkageStanding}.
This picture depicts a four-bar linkage standing on two flat, rigid contacts. In this configuration, the mechanical system possesses one degree of freedom, i.e. the variable $\xi$, which represents the minimal coordinate of the constrained mechanical system. At the equilibrium configuration, the system is \emph{hyperstatic} because the equilibrium conditions of the Newton-Euler equations 
are not sufficient to determine all the (internal) joint torques $\tau$. 
Since the joint torques are assumed to be a control input, however, one can assume that their value is given, with the only requirement that the system remains at the equilibrium configuration.

Assume that the joint torques depend only on the minimal coordinate $\xi$. Then, the center of pressures at each \emph{static} robot configuration, i.e. $\dot{\xi}=0$, depend only on the minimal coordinate~$\xi$. This poses the following interesting question.

$\\ $
\noindent
    \emph{What is the rate-of-change of the center of pressures over the quasi-static constrained robot's motion?}
$\\ $

Different control laws for balancing the four-bar linkage may be associated with different rate-of-changes of the static center of pressure over the constrained robot's motion. Clearly, the higher this rate-of-change, the higher the likelihood to hit the boundaries of the support  polygon of the foot, which causes breaking the associated contact and, eventually, a fall of the robot. The above rate-of-change of the center of pressure is what we call \emph{static center of pressure sensitivity}, and the following generalises this concept to any multi-body system constrained by some rigid, flat contacts.

\subsection{The formal definition}
Assume that the robot makes $n_c$ rigid contacts with the environment, and that no external wrench acts on the robot other than gravity and contact wrenches. Let ${J}$ denote the Jacobian obtained by stacking all contact Jacobians ${J}_{\mathcal{C}_k}$, i.e.
\begin{IEEEeqnarray}{RCL}
	\label{eqn:jacobian_formal}
	{J} = 
	\begin{bmatrix}
	{J}_{\mathcal{C}_1} \\
	\vdots \\
	{J}_{\mathcal{C}_{n_c}}
	\end{bmatrix} \in \mathbb{R}^{k \times n+6},
\end{IEEEeqnarray}
with $k = 6 n_c$.
Consequently, all rigid contacts imply the following constraint on the floating base system~\eqref{eq:system_dynamics}: ${J} {\nu} = 0$,
the time derivative of which is given by
\begin{IEEEeqnarray}{RCL}
	\label{eqn:completeConstraintAcc}
	\dot{{J}} {\nu} + {{J}}\dot{{\nu}} = 0.
\end{IEEEeqnarray}
By combining the above constraint with Eq.~\eqref{eq:system_dynamics} and by defining 
\[f := ({f}^\top_1,\dots,{f}^\top_{n_c})^\top \in \mathbb{R}^{k},\] 
one can find the explicit expression of all contact forces, i.e.
\begin{align}
    \label{eq:contact_forces}
    f= ({J} {M}^{-1}{J}^\top)^{-1}\left[ {J} {M}^{-1} \big({C}(q,{\nu}){\nu} {+} {G}(q) {-} B\tau \big) {-} \dot{{J}}{\nu} \right].
\end{align}
The hidden assumption for the above equality to hold is that~${J}$ is full row rank. 
Once the obtained expression of $f$ is substituted into Eq.~\eqref{eq:system_dynamics}, 
one obtains the following equations of motion:
\begin{IEEEeqnarray}{rCl}
    {M}(q)\dot{{\nu}} &+& {J}^\top ({J}M^{-1}{J}^\top)^{-1} \dot{{J}}{\nu}   = N \left[ B \tau - {C}(q, {\nu}) {\nu} - {G}(q) \right] 
    \IEEEeqnarraynumspace 
\label{eq:forwardWithCon}
\end{IEEEeqnarray}
with $N$ given by
\begin{IEEEeqnarray}{rCl}
    N &=& 1_{n+6} - {J}^\top ({J} {M}^{-1}{J}^\top)^{-1} {J} {M}^{-1}. \notag 
\end{IEEEeqnarray}
Evaluating Eq.~\eqref{eq:forwardWithCon} at the equilibrium condition 
$({\nu},\dot{{\nu}})=(0,0)$ yields
\begin{IEEEeqnarray}{rcl}
    N \left[ B \tau - {G}(q) \right] =0.
    \label{eq:tauEquilibrium}
\end{IEEEeqnarray}
Eq.~\eqref{eq:tauEquilibrium} defines the values that the joint torques must take to satisfy the equilibrium condition $({\nu},\dot{{\nu}})=(0,0)$. Then, we make the following assumption.

\begin{assumption} 
\label{hp:constrainedMotionAndEqTorque}
The holonomic constraints due to contacts restrain the configuration space $\mathbb{Q}$ into the set
$\mathbb{R}^{n+6-k}$, i.e. there exists a differentiable mapping 
$\chi : \mathbb{R}^{n+6-k} \rightarrow \mathbb{Q}  $ such that 
\begin{IEEEeqnarray}{rcl}
    \forall q \in \mathbb{Q} \quad \exists~ \xi \in \mathbb{R}^{n+6-k} \quad : \quad q = \chi(\xi).
    \label{eqn:mappinMinimulCoorConfSpa}
\end{IEEEeqnarray}
Furthermore, the joint torques $\tau$ at the equilibrium configuration
$({\nu},\dot{{\nu}})=(0,0)$ 
depend only on the robot constrained configuration space, i.e. $\tau = \tau(\xi)$. 
\end{assumption}

In view of the above assumption, the contact wrenches at the equilibrium configuration, i.e. 
\begin{IEEEeqnarray}{RCL}
    \label{eq:contactforcesEq}
    f^e= 
	\begin{bmatrix}
	f^e_1 \\
	\vdots \\
	f^e_{n_c}
	\end{bmatrix} =   
    ({J} {M}^{-1}{J}^\top)^{-1} {J}M^{-1} \left( {G} {-} B\tau \right),
\end{IEEEeqnarray}
 depend only on the robot constrained motion, i.e. $f^e= f^e(\xi)$.
In light of the above, we can now define the \emph{static center of pressure sensitivity} as follows.
\begin{definition}
\label{def:scopSens}
Let $\text{SCoP}_j \in \mathbb{R}^2$ denote the  center of pressure of the $j$-th rigid contact at the equilibrium $({\nu},\dot{{\nu}})=(0,0)$, i.e.
\begin{IEEEeqnarray}{RCL}
    \label{eq:scop}
    \text{SCoP}_j  = \frac{1}{e_3^\top f^e_j} \left.
    \begin{bmatrix}
	 -e_5^\top f^e_j \\
	 e_4^\top f^e_j
    \end{bmatrix} \right. .
\end{IEEEeqnarray}
The \emph{static center of pressure sensitivity} $\eta_j$ is defined as the rate-of-change of~\eqref{eq:scop}  over the  constrained robot motion, i.e. 
\begin{IEEEeqnarray}{RCL}
    \label{eq:eta}
    \eta_j : = \frac{\partial~ \text{SCoP}_j }{\partial \xi}.
\end{IEEEeqnarray}
\end{definition}

In general, the \emph{static center of pressure sensitivity} is a matrix belonging to 
$\mathbb{R}^{2\times n+6-k}$, and characterises how fast the center of pressure associated with a rigid, planar contact moves depending on the constrained robot's motion.
This sensitivity may then characterise ``the stability'' of a contact associated with a robot's configuration. 
Of particular importance are the cases when the sensitivity $\eta_j$ tends to infinity. 
Clearly, these configurations must be avoided by any controller associated with the robot, since they may cause abrupt system's responses and, eventually, a robot fall.
\begin{Remark}
The internal torques at the equilibrium configuration must satisfy Eq. \eqref{eq:tauEquilibrium} that induces, in general, multiple solutions when solved for $\tau$. The proposed sensitivity may be then used to compare different balancing controllers, since lower values of $\eta_j$ are associated with smaller variations of the center of pressures, which may increase the likelihood of not breaking the contact.  
\end{Remark}
\begin{Remark}
    Note that, by its definition in Eq. \eqref{eq:eta}, the actual value of the sensitivity depends on the choice of the parametrisation $\xi$. Different choices of parametrisation may lead to different values of the sensitivity.
\end{Remark}

\section{The sensitivity for assessing balancing controllers performance:\\ the four-bar linkage case study}
\label{sec:four-bar}
This section studies how the sensitivity of the static center of pressure may vary depending on different balancing controllers synthesised for the same system.
What follows presents the analytical analysis on a four-bar linkage when standing on two, flat, rigid contacts.
For balancing purposes of the four-bar linkage, 
we choose two state-of-the-art momentum-based controllers that differ in the criterion adopted for the contact forces redundancy resolution.
More precisely,
we show next that minimising the joint torques when the Newton-Euler equations are at the equilibrium decreases the \emph{static center of pressure sensitivity}, thus improving the stability of the contact.
As the experimental results in Section~\ref{sec:experiments} show, the four-bar linkage case study  is representative of a humanoid robot standing on two feet.
\subsection{Modelling}
\begin{figure}[t]
    \centering{
        \def\svgwidth{\columnwidth}
        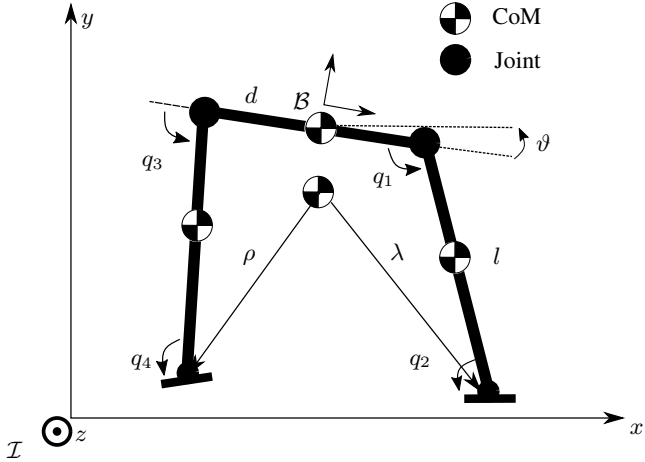
    \caption{Free floating planar four-bar linkage}
    \label{fig:simple_model}
    }
\end{figure}

\begin{figure*}[t]
  \centering
    \subfloat[Minimum wrenches norm solution]{\includegraphics[width=.9\columnwidth]{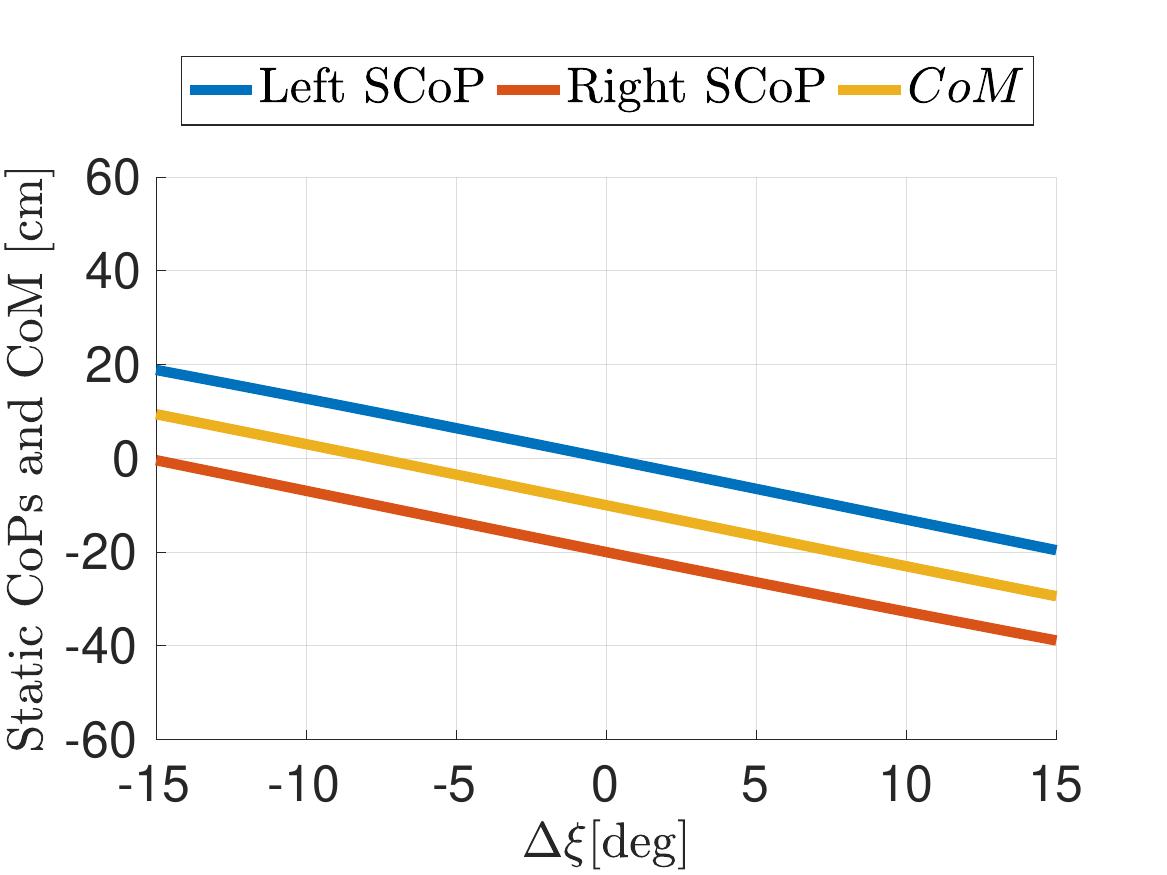}\label{fig:simple_minforces_cop}}
    \subfloat[Minimum joint torque norm solution]{ \includegraphics[width=.9\columnwidth]{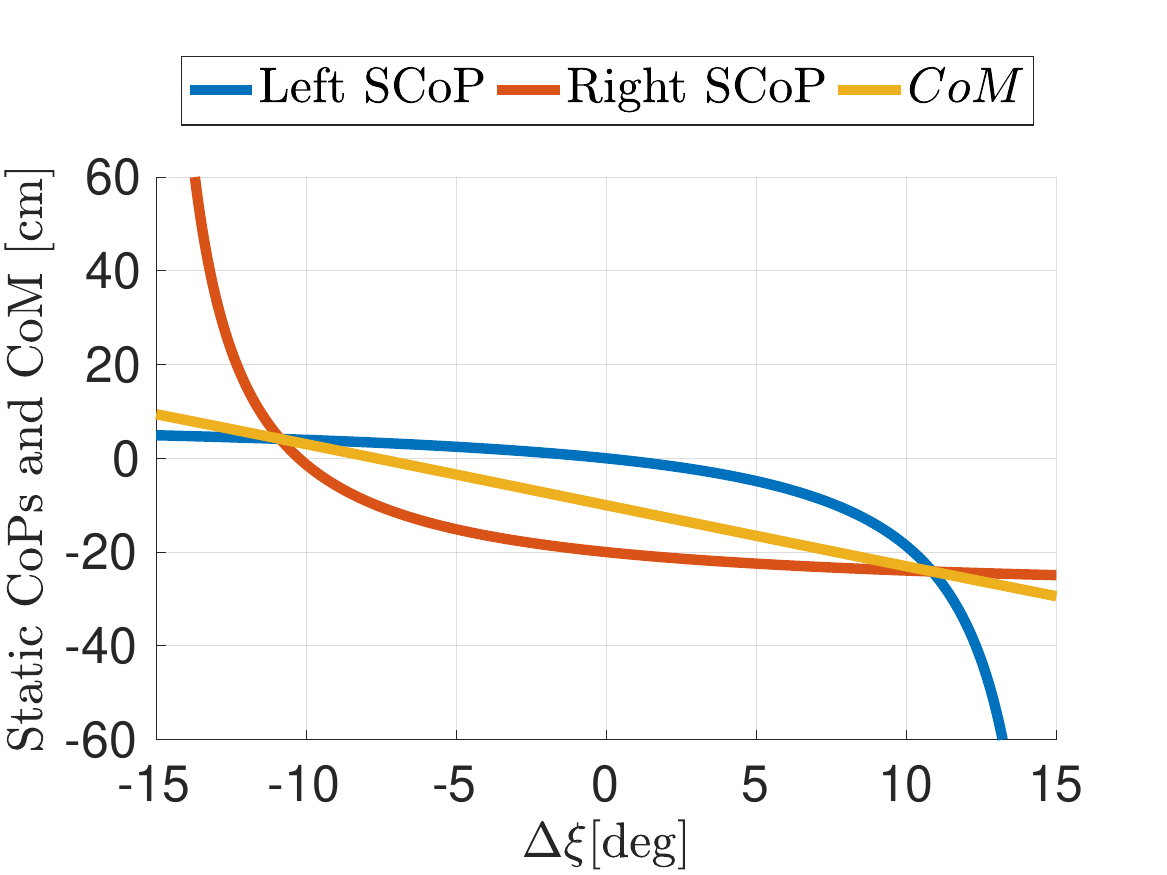}\label{fig:simple_mintorques_cop}}
    \label{fig:forces_4bar}
    \caption{Left and right foot CoPs together with the CoM for different configuration of $\xi$. In \protect \subref{fig:simple_mintorques_cop} the asymptotes are in correspondence of the vertical component of the forces going to zero
    }
\end{figure*}

\begin{figure}[t]
  \centering
  \includegraphics[width=.9\columnwidth]{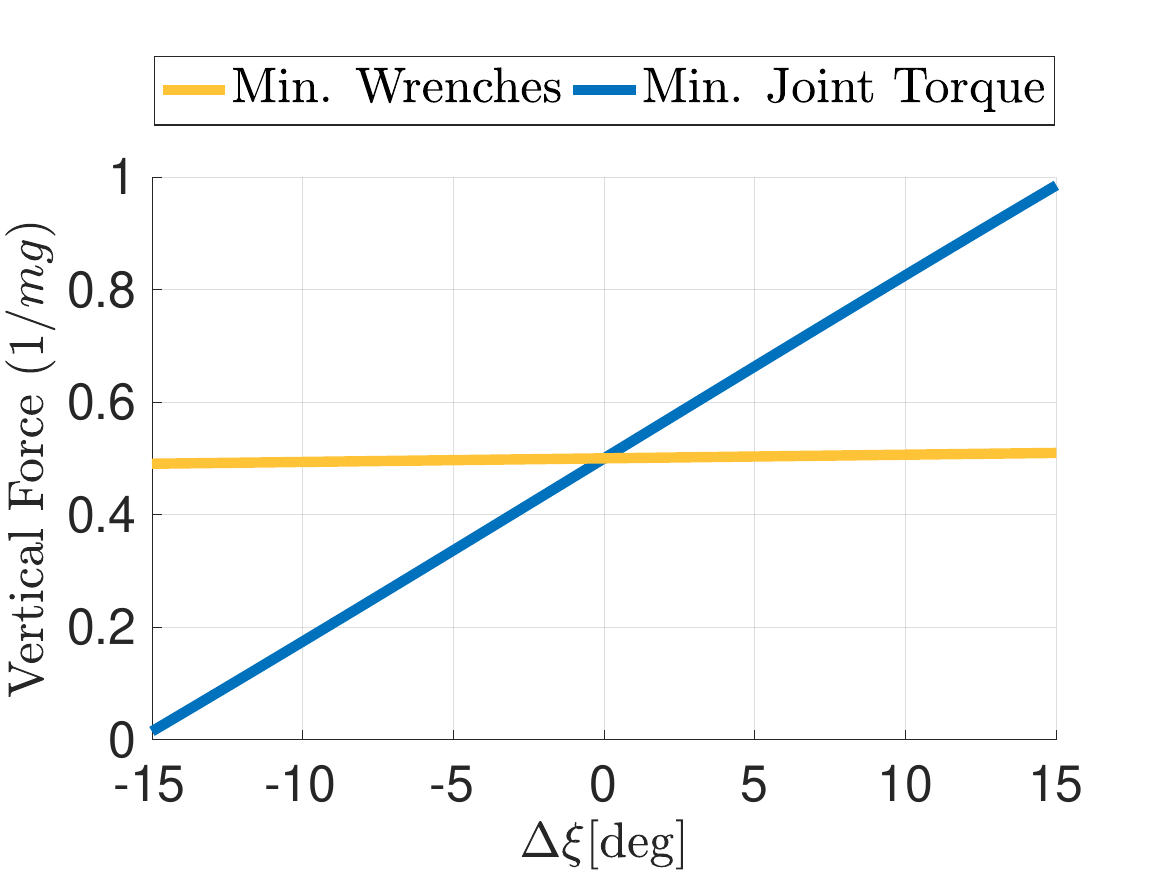}
  \caption{Comparison of the vertical force component of the right foot for the two contact forces choice. The vertical force is normalised w.r.t. the total mass of the mechanism, i.e. divided by $1/mg$
  }
    \label{fig:forces_comparison_rightfoot}
\end{figure}

Consider the  four-bar linkage shown in Figure~\ref{fig:simple_model}, where $l$ and 
$d$ denote the lengths of the leg and of the upper rod, respectively. 
Each leg possesses a mass of $m_l$ while the upper rod has mass $m_b$.
The inertial frame $\mathcal{I}$ is chosen so as the $y$ axis opposes gravity, the $z$ axis exits from the plane, and the $x$ axis completes the right-hand base.
Being a planar model,  only translations in the $x-y$ plane and rotations about the $z$ axis are allowed. 
The vectors 
\[\lambda = \begin{bmatrix}
\lambda_1 & \lambda_2
\end{bmatrix}^\top \in \mathbb{R}^2,\quad \rho = \begin{bmatrix}
\rho_1 & \rho_2
\end{bmatrix}^\top \in \mathbb{R}^2\] connect the center of mass to the left and right foot frames respectively. 
Let $^\mathcal{I}{p}_\mathcal{B} \in \mathbb{R}^2$ denote the position of the origin of the frame~$\mathcal{B}$ -- which is located at the center of the upper link -- expressed in the inertial frame. The configuration space is then parametrised by the following coordinates 
\[q = (^\mathcal{I}{p}_\mathcal{B},\theta, q_1,q_2,q_3,q_4) \in \mathbb{R}^7.\]
Being a vector space, we use the Lagrange formalism to derive the equations of motion \cite[Ch. 4]{Sciavicco2nd}, i.e.
\begin{equation}
    \label{eq:lagrangian}
    \frac{\dif}{\dif t} \frac{\partial L}{\partial \dot{q}} + \frac{\partial L}{\partial q}
     = \begin{bmatrix}0_3 \\ \tau\end{bmatrix},
\end{equation}
with $L := T - V$ the Lagrangian, and $T$ and $V$ respectively the kinetic and potential energy of the multi-body system.
\begin{hypothesis}
    \label{hyp:four_bar}
    For the four-bar linkage model, it holds that:
    \begin{enumerate}[i)]
        \item Each link is approximated as a point mass at its center of mass.
        \item The mass of the left leg equals the mass of the right leg.
        \item The upper body mass, $m_b$, is twice the leg mass, i.e. $m_b = 2 m_l$.
    \end{enumerate}
\end{hypothesis}
    It is worth noting that hypothesis \emph{ii)} and \emph{iii)} are in general satisfied in  humanoid robots, e.g. in the iCub robot used for the experiments. 
    
For the sake of simplicity, the complete dynamic model obtained by applying Eq. \eqref{eq:lagrangian} is not presented here. 
We present only the terms of the dynamic and kinematic model that are necessary for the comprehension of the remainder of the section. In particular, partition the mass matrix $\xoverline{M}$ as follows 
\begin{IEEEeqnarray}{RCL}
\xoverline{M} &=& 
\begin{bmatrix}
	{M}_b && {M}_{bj} \\
	{M}^\top_{bj} && {M}_{j}
\end{bmatrix}.
\nonumber
\end{IEEEeqnarray}
Let $\sin(q_i) = s_i$ and $\cos(q_i) = c_i$. One can verify that 
\begin{IEEEeqnarray}{RCL}
\label{MbjFourBarLink}
{M}_{bj} &=& \frac{m_l}{2}
\begin{bmatrix}
	l s_1 & 0 & l s_3 & 0 \\
	- l c_1 & 0 & - l c_3 & 0 \\
	 \frac{l^2 s^2_1 {-} l c_1(d{-}l c1)}{2} & 0 &  \frac{l^2 s^2_3 {+} l c_3(d{+}l c_3)}{2} & 0 
\end{bmatrix} \IEEEeqnarraynumspace
\end{IEEEeqnarray}
The feet Jacobians are given by the following expression
\begin{equation*}
    \label{eq:fourbarJacobian}
    \begin{aligned}
        {J}_{\mathcal{C}_L} &=
        \left[\begin{array}{c c c c c c c}
             &  &  &  & 0 & 0 & 0 \\
            \multicolumn{2}{c}{\smash{\raisebox{.5\normalbaselineskip}{$1_2$}}} &
            \smash{\raisebox{.5\normalbaselineskip}{$R(\theta)\begin{bmatrix}
                ls_1\\ \frac{d}{2} -l c_1
            \end{bmatrix}$}} 
            & \smash{\raisebox{.5\normalbaselineskip}{$R(\theta)\begin{bmatrix}
                l  s_1\\
                -l  c_1
            \end{bmatrix}$}} & 0 & 0 & 0 \\
            0 & 0 & 1 & 1 & 1 & 0 & 0
        \end{array}\right]\\
        {J}_{\mathcal{C}_R} &=
        \left[\begin{array}{c c c c c c c}
             &  &  &   0 & 0 & & 0 \\
            \multicolumn{2}{c}{\smash{\raisebox{.5\normalbaselineskip}{$1_2$}}} &
            \smash{\raisebox{.5\normalbaselineskip}{$R(\theta) \begin{bmatrix}
                ls_3\\ -\frac{d}{2} +l c_3
            \end{bmatrix}$}} 
            & 0 & 0 & \smash{\raisebox{.5\normalbaselineskip}{$R(\theta) \begin{bmatrix}
                l s_3 \\ -l c_3
            \end{bmatrix}$}}  & 0 \\
            0 & 0 & 1 & 0 & 0 & 1 & 1
        \end{array}\right].
    \end{aligned}
\end{equation*}

To represent a classical balancing context of humanoid robots, we assume that the mechanism makes contact with the environment at the two extremities. Also, we assume  that 
the distance between the contacts equals $d$, i.e. the length of the upper rod -- see Figure \ref{fig:fourBarLinkageStanding}. Then, the rigid constraint acting on the system takes the form~\eqref{eqn:completeConstraintAcc}, with 
${J} = 
\begin{bmatrix}
{J}_{\mathcal{C}_L}^\top  &
{J}_{\mathcal{C}_R}^\top
\end{bmatrix}^\top \in \mathbb{R}^{6 \times 7}
$,
and
$\nu = \dot{q}$. The associated contact wrenches are $f = (f_L,f_R) \in \mathbb{R}^6$.
As a consequence of these constraints, the mechanism  possesses one degree of freedom when standing on the extremities.

With the aim of evaluating the sensitivity of the static center of pressure we define the minimal coordinate $\xi$ as the angle between the upper link and one of the legs (see Figure \ref{fig:fourBarLinkageStanding}).
Observe that the mapping $q = \chi(\xi)$ in~\eqref{eqn:mappinMinimulCoorConfSpa} is given by:
\begin{IEEEeqnarray}{RCL}
\begin{bmatrix}
^\mathcal{I}{p}_b \\
\theta \\ 
 q_1 \\
 q_2 \\ 
 q_3 \\
 q_4 \\
\end{bmatrix} &=&
\begin{bmatrix}
\tfrac{1}{2}d e_1 + l \sin(\xi) e_2+ l \cos(\xi) e_1 \\
0 \\ 
 \xi \\
\pi -\xi \\ 
 \xi \\
 \pi-\xi \\
\end{bmatrix}.
\end{IEEEeqnarray}

\subsection{Two balancing controllers inducing different sensitivities}\label{sec:choice_of_wrenches}

Remark that in order to evaluate the sensitivity of the static center of pressure, we need an expression for the contact forces at the equilibrium configuration -- see Definition~\ref{def:scopSens}.
What follows proposes an analysis of the sensitivity of the static center of pressures evaluated with contact wrenches obtained from two state-of-the-art criteria.
\begin{enumerate}
\item The contact wrenches ensure the equilibrium of the centroidal dynamics written in the plane.
The three dimensional redundancy of the contact wrenches minimises the norm of the contact wrenches\footnote{It is worth noting that the norm of a wrench is not well defined and highly dependent on the chosen metric. Nevertheless, the minimisation of the contact wrench norm is something extensively adopted in literature and it has been used in the manuscript only for comparison purposes.} $|f|$ \cite{Herzog2014, Hopkins2015b,Lee2010,Liu2015,Nori2015}.
\item The contact wrenches ensure the equilibrium of the centroidal dynamics.
The three dimensional redundancy of the contact wrenches minimises the norm of the joint torques $|\tau|$ \cite{Nava16,PucciVideo2016}.
\end{enumerate}

The following lemma presents the results when the first criterion is applied.
\begin{boldLemma}
    \label{lemma:min_wrench}
Assume that Hypothesis \ref{hyp:four_bar} holds. If the contact wrenches redundancy is chosen to minimise the norm of the contact wrenches, i.e. criterion $1)$, then the induced contact wrenches are given by the following expression:
\begin{equation}
    \label{eq:minforces_wrenches}
    \begin{aligned}
    \prescript{L}{}f_L &=
    {mg}
    \begin{bmatrix}
         0\\[0.5em]
         \frac{1}{2 \left(1 + \frac{d^2}{4}\right)} + d\frac{5 d - 6l \cos(\xi)}{10 (d^2 + 4)} \\[0.5em]
         \frac{6l \cos(\xi)}{5 (d^2 + 4)}
    \end{bmatrix},\\
    \prescript{R}{}f_R &=
    {mg}
    \begin{bmatrix}
         0\\[0.5em]
         \frac{1}{2 \left(1 + \frac{d^2}{4} \right)} + d\frac{5 d + 6l \cos(\xi)}{10 (d^2 + 4)} \\[0.5em]
         \frac{6l \cos(\xi)}{5 (d^2 + 4)}
    \end{bmatrix}.
    \end{aligned}
\end{equation}
Furthermore, the static center of pressure sensitivity is given by:
\begin{equation}
    \label{eq:minforces_cop_sens}
   \begin{aligned}
       \eta_L &= -60 l \sin(\xi) \frac{d^2 + 4}{(5 d^2 + 6 l d \cos(\xi) + 20)^2},\\
       \eta_R &= -60 l \sin(\xi) \frac{d^2 + 4}{(5 d^2 - 6 l d \cos(\xi) + 20)^2}.
   \end{aligned}
\end{equation}
\end{boldLemma}
The proof is in the Appendix.

It is worth noting that if $d$ is small enough,
then Eq. \eqref{eq:minforces_wrenches} points out that one has an \emph{almost} constant vertical force independently of $\xi$.
In this case, the vertical force is approximately equal to $mg/2$, independently of the center-of-mass position.

Figure~\ref{fig:simple_minforces_cop} shows the static CoP, referred to as \emph{SCoP}, of the left and right foot in correspondence of the CoM motion. We can notice how $\text{SCoP}_L$ (the SCoP of the left foot) and $\text{SCoP}_R$ (the SCoP of the right foot) moves together with the CoM.

Let us now present analogous results of Lemma~\ref{lemma:min_wrench} when applying the criterion 2), i.e. the choice of the force redundancy minimises the joint torques.

\begin{boldLemma}
    \label{lemma:min_torques}
    Assume that Hypothesis \ref{hyp:four_bar} holds. 
    If the contact wrenches redundancy is chosen to minimise the norm of the internal joints torques, i.e. criterion $2)$, the induced wrenches are given by the following expressions.
    \begin{subequations}
     \label{eq:mintorques_wrenches}
         \begin{equation}
             \label{eq:mintorques_wrenches_fL}
            {}^L f_L {=} mg \left( \begin{bmatrix}
                0 \\ \frac{1}{2} \\ -\frac{\lambda_1}{2}
            \end{bmatrix} {+} \begin{bmatrix}
                \frac{3l\cos(\xi)^2} {8 d \sin({\xi})} \\
                \frac{3l \cos({\xi})}{8d} \\
                \frac{3l \cos({\xi})}{8d} \left( \lambda_2 \frac{\cos(\xi)}{\sin({\xi})} - \lambda_1 \right) + \frac{d}{4}
            \end{bmatrix} \right),
         \end{equation}
         \begin{equation}
                  \label{eq:mintorques_wrenches_fR}
             {}^R f_R {=} mg \left( \begin{bmatrix}
                 0 \\ \frac{1}{2} \\ -\frac{\rho_1}{2}
             \end{bmatrix} {-} \begin{bmatrix}
                 \frac{3l\cos(\xi)^2} {8 d \sin({\xi})} \\
                 \frac{3l \cos({\xi})}{8d} \\
                 \frac{3l \cos({\xi})}{8d} \left( \rho_2 \frac{\cos(\xi)}{\sin({\xi})} - \rho_1 \right) + \frac{d}{4}
             \end{bmatrix} \right).
         \end{equation}
     \end{subequations}
     
     Furthermore, the static center of pressure sensitivity is given by:
     \begin{equation}    
         \label{eq:mintorques_cop_sens}
        \begin{aligned}
    \eta_L &= - \frac{45d^2l\sin(\xi)}{(10d + 3l\cos(\xi))^2},\\
   \eta_R &= - \frac{45d^2l\sin(\xi)}{(10d - 3l\cos(\xi))^2}.
        \end{aligned}
     \end{equation}
\end{boldLemma}
The general expression of the contact wrenches \eqref{eq:mintorques_wrenches} when Hypothesis \ref{hyp:four_bar}-\emph{iii)} does not hold is given in the Appendix.

Focus on the contact wrenches in \eqref{eq:mintorques_wrenches}. 
Differently from Eq.~\eqref{eq:minforces_wrenches}, the vertical components of the contact forces ${}^L f_L$ and ${}^L f_R$ in equations \eqref{eq:mintorques_wrenches_fL} and \eqref{eq:mintorques_wrenches_fR} do vary with the variable $\xi$, see Figure \ref{fig:forces_comparison_rightfoot}.
In particular, when the center of mass of the system is on top of one of the feet, the corresponding vertical force is equal to $mg$, i.e. the foot supports the whole weight. As a consequence, the force on the other foot is null.

\begin{figure}[t]
  \centering
 \includegraphics[width=.9\columnwidth]{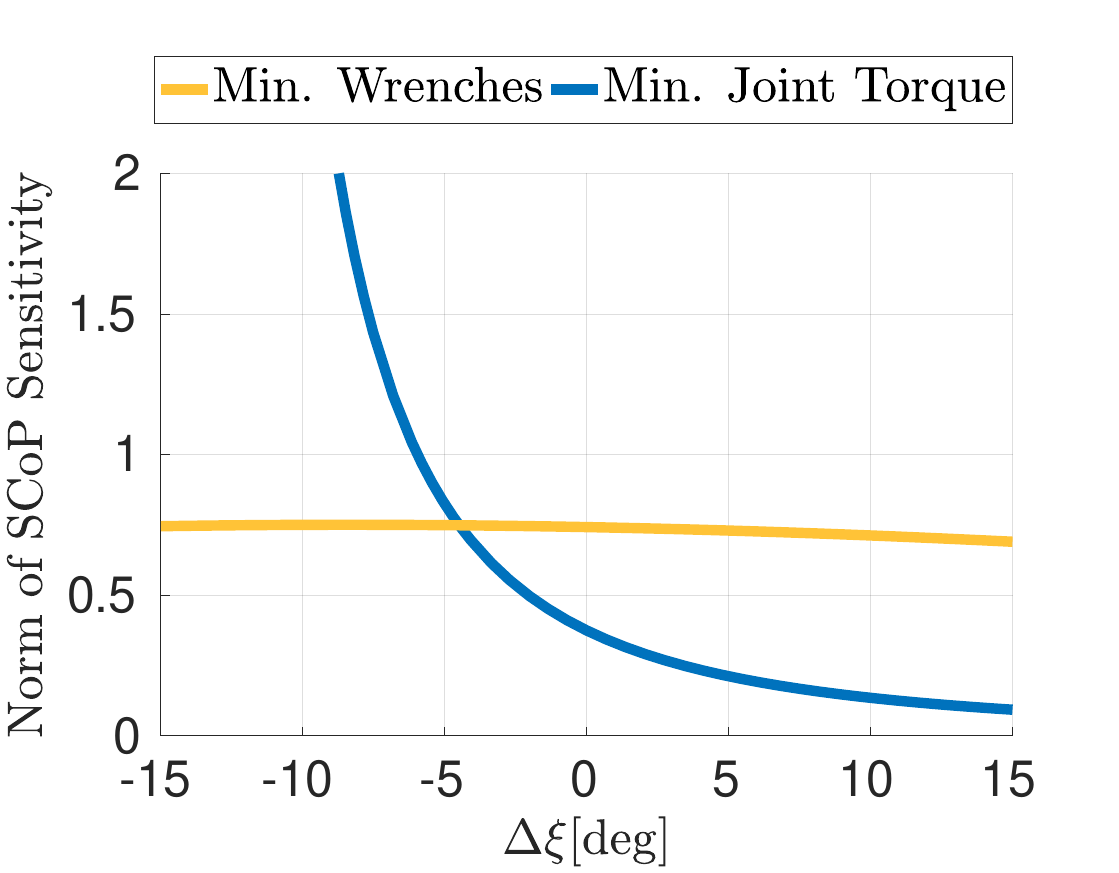}
  \caption{Comparison of the norm of the sensitivity of the static center of pressure of the right foot for the two contact forces choice
  }
  \label{fig:sensitivity}
\end{figure}

The behaviour of the static centers of pressure can be observed in Figure~\ref{fig:simple_mintorques_cop}.
Interestingly, while the minimum torques norm solution exhibits more ``flat'' CoPs around the centered position and on the support foot, we can notice a divergent behaviour around $\Delta \xi= \pm 14 [\mathrm{deg}]$.
This is due to the behaviour of the vertical components of the force previously described, i.e. to the fact that with this solution, the force on one foot tends to zero as the weight is moved towards the other foot.
It is worth noting that the center of pressure showing the divergent behaviour is the one of the non-supporting foot.


We can compare the sensitivity of the static centers of pressure induced by criteria 1) and 2) when the CoM moves from left to right.
Figure~\ref{fig:sensitivity} compares the norm of the right foot static center of pressure sensitivities obtained by applying the two criteria.
Notice that the second criterion, i.e. the minimum joint torques norm solution, provides less sensitive solutions at the initial, symmetrical configuration, i.e. $\Delta \xi = 0$.
Furthermore, the sensitivity decreases when the foot is increasingly supporting all the weight, i.e. toward $\Delta \xi = 14[ \mathrm{deg}]$ in the considered case.

\vspace{0.1cm}
\begin{Remark}
A criterion similar to the minimisation of the norm of the contact wrenches that is sometimes found in literature is the minimisation of the tangential components of the forces \cite{Righetti01032013}.
The following lemma states the equivalence of the two solutions when planar contacts are considered. The proof can be found in the Appendix.
\end{Remark}

\begin{boldLemma}
    \label{lemma:min_tang}
    The solution exploiting the force redundancy to minimise the tangential component of the contact wrenches is equivalent to the minimum contact wrench norm solution (Lemma \ref{lemma:min_wrench}).
\end{boldLemma}

\begin{Remark} The proposed analysis focuses on the sensitivity of the center of pressure, that is, on its rate of change with respect to a parametrisation of the minimal coordinates of the constrained system.
The use of inequalities on the contact wrenches, usually enforced by the use of QP solvers, limits the actual value of the center of pressure.
The two concepts are thus complementary. 
\end{Remark}
\section{Experiments}
\label{sec:experiments}


\begin{figure*}[t]
  \centering
    \subfloat[Minimum wrenches norm solution]{\includegraphics[width=.9\columnwidth]{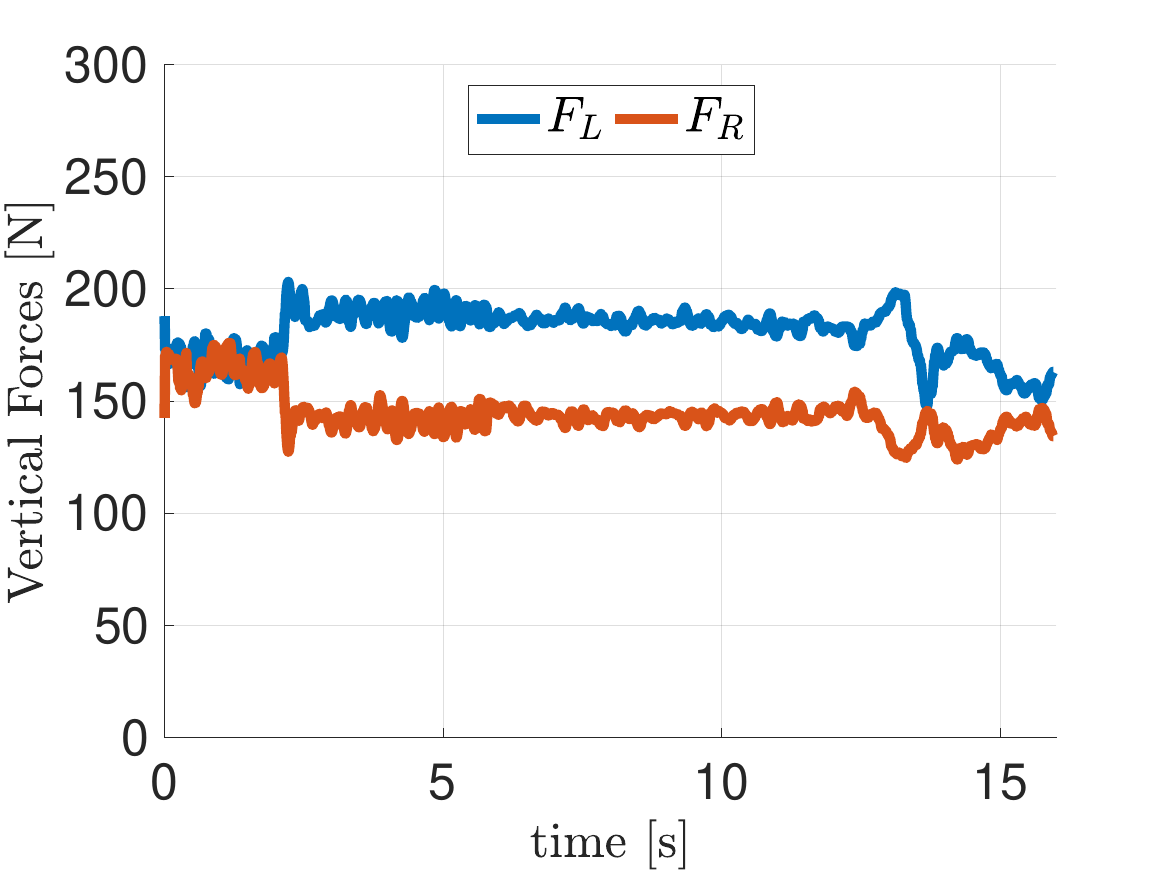}\label{fig:robot_minforces_forces}}
    \subfloat[Minimum joint torque norm solution]{ \includegraphics[width=.9\columnwidth]{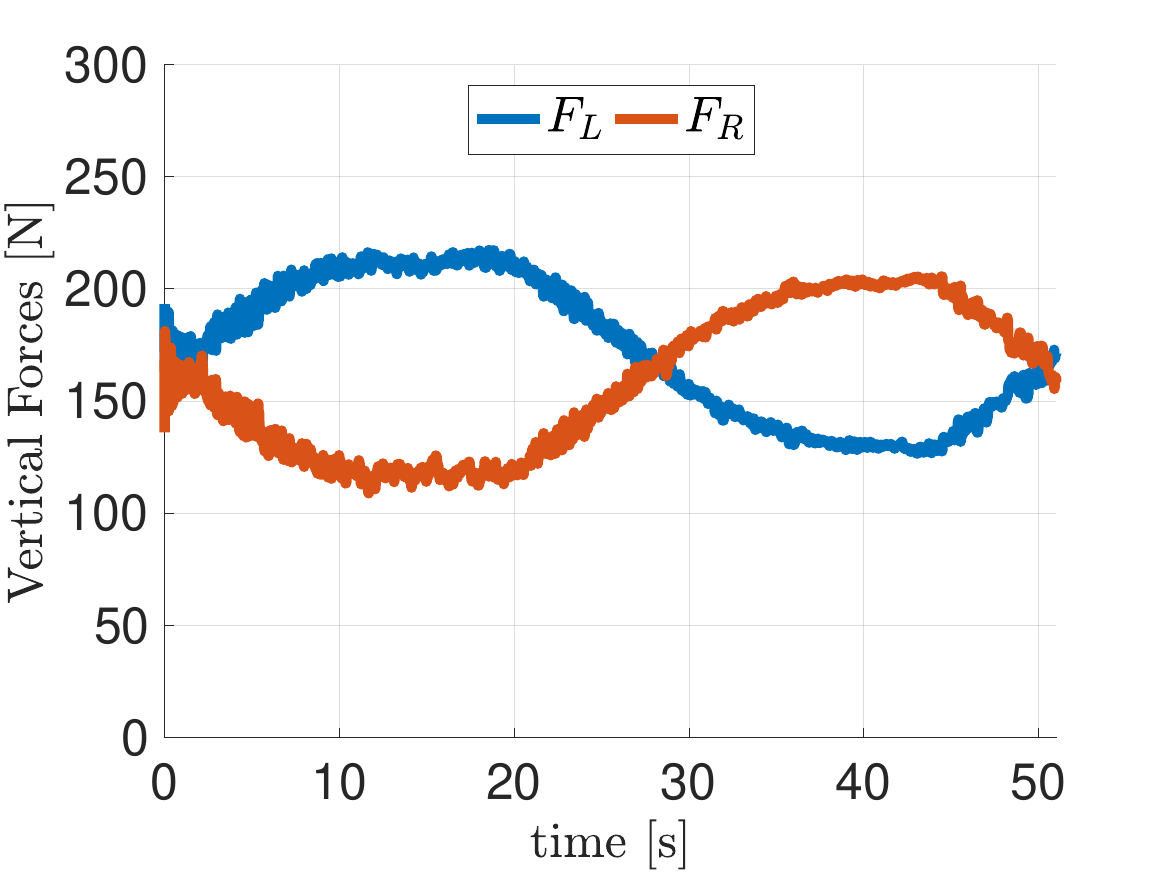}\label{fig:robot_mintorques_forces}}
    \label{fig:forces_robot}
    \caption{Vertical components of the contact forces for left and right foot during the robot quasi-static motion. The experiment in \protect\subref{fig:robot_minforces_forces} depicts the forces when the CoM moves towards the left foot. In \protect\subref{fig:robot_mintorques_forces} the CoM tracks a full sinusoidal reference period instead.}
\end{figure*}

\begin{figure*}[t]
  \centering
    \subfloat[CoM towards left foot]{\includegraphics[width=.9\columnwidth]{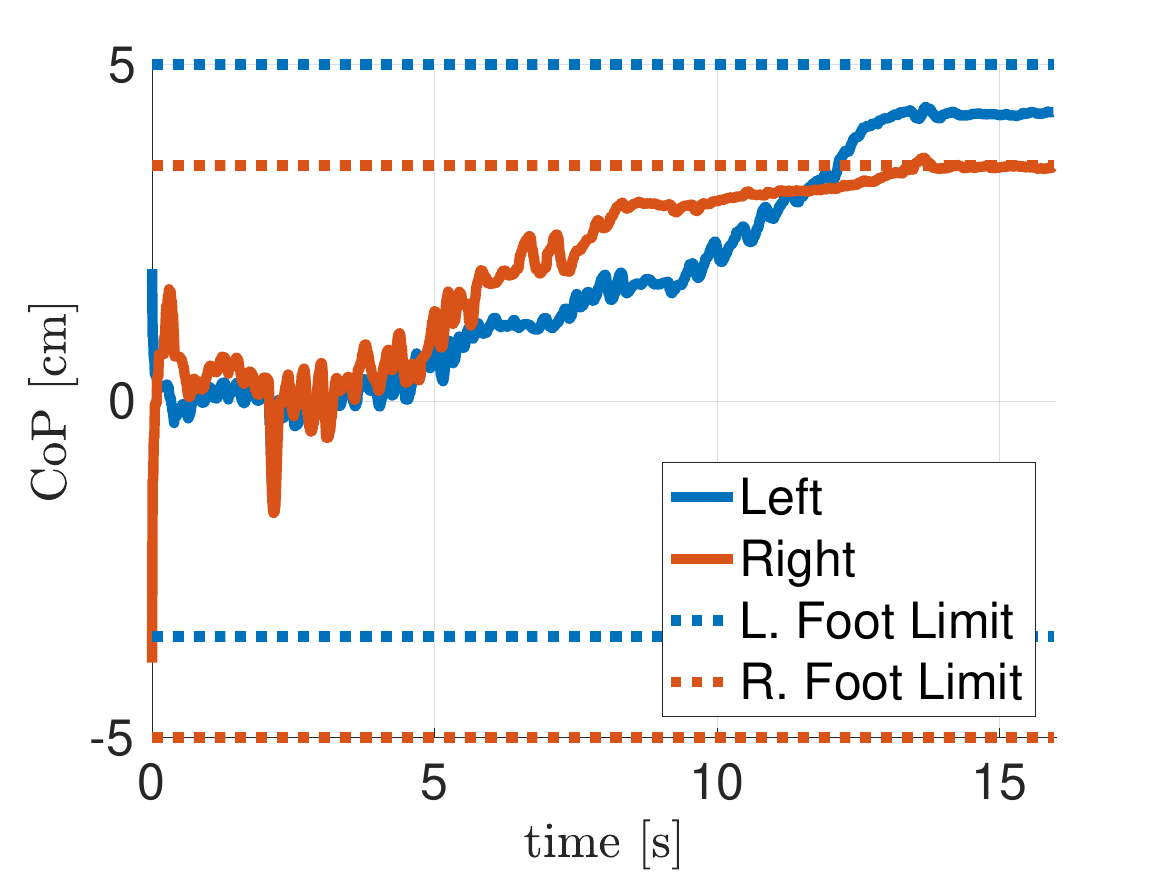}\label{fig:cop_min_wrenches_leftcom}}
    \subfloat[CoM towards right foot]{ \includegraphics[width=.9\columnwidth]{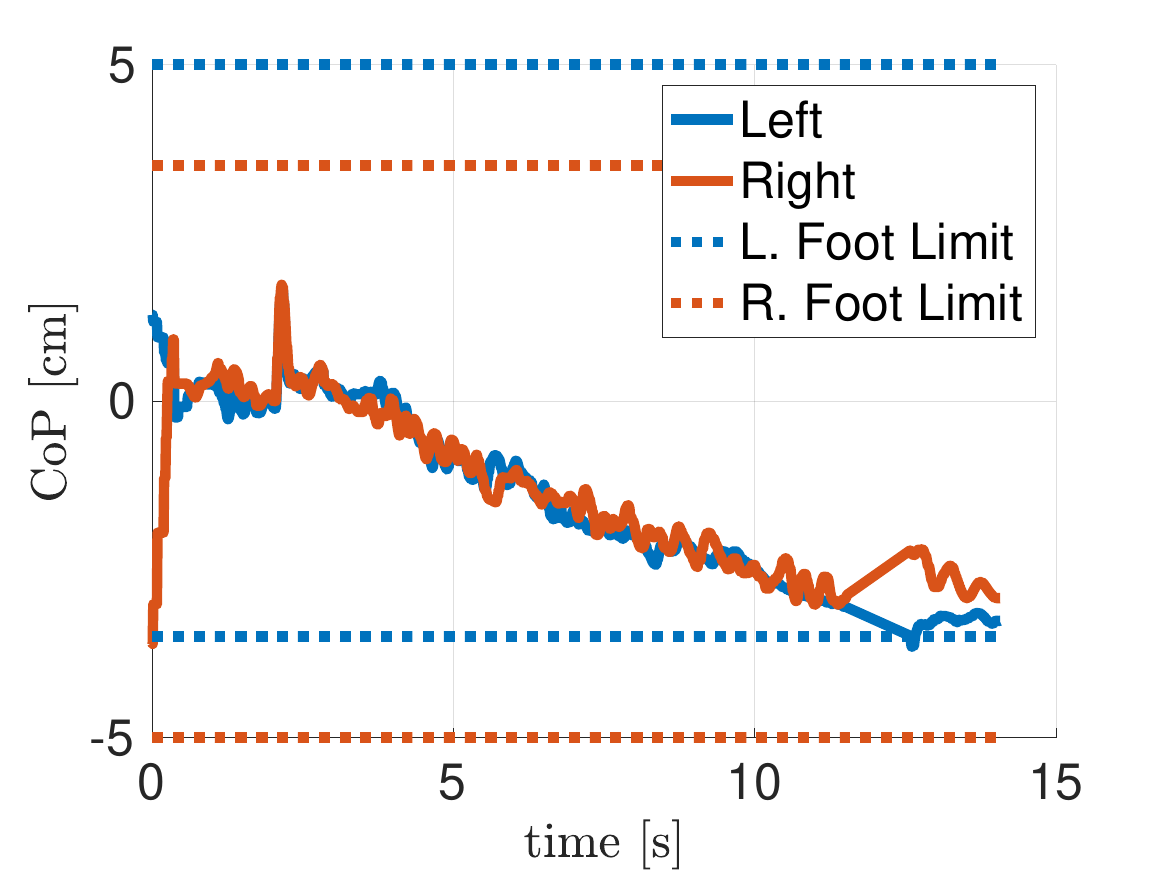}\label{fig:cop_min_wrenches_rightcom}}
    \caption{Center of pressure for the left and right foot when the minimisation of the contact wrenches criterion is applied. In \protect\subref{fig:cop_min_wrenches_leftcom} we show the experiment with the CoM moving towards the left foot, while in \protect\subref{fig:cop_min_wrenches_rightcom} the CoM moves towards the right foot instead.}
    \label{fig:cop_min_wrenches}
\end{figure*}

\begin{figure}[t]
  \centering
  \includegraphics[width=.9\columnwidth]{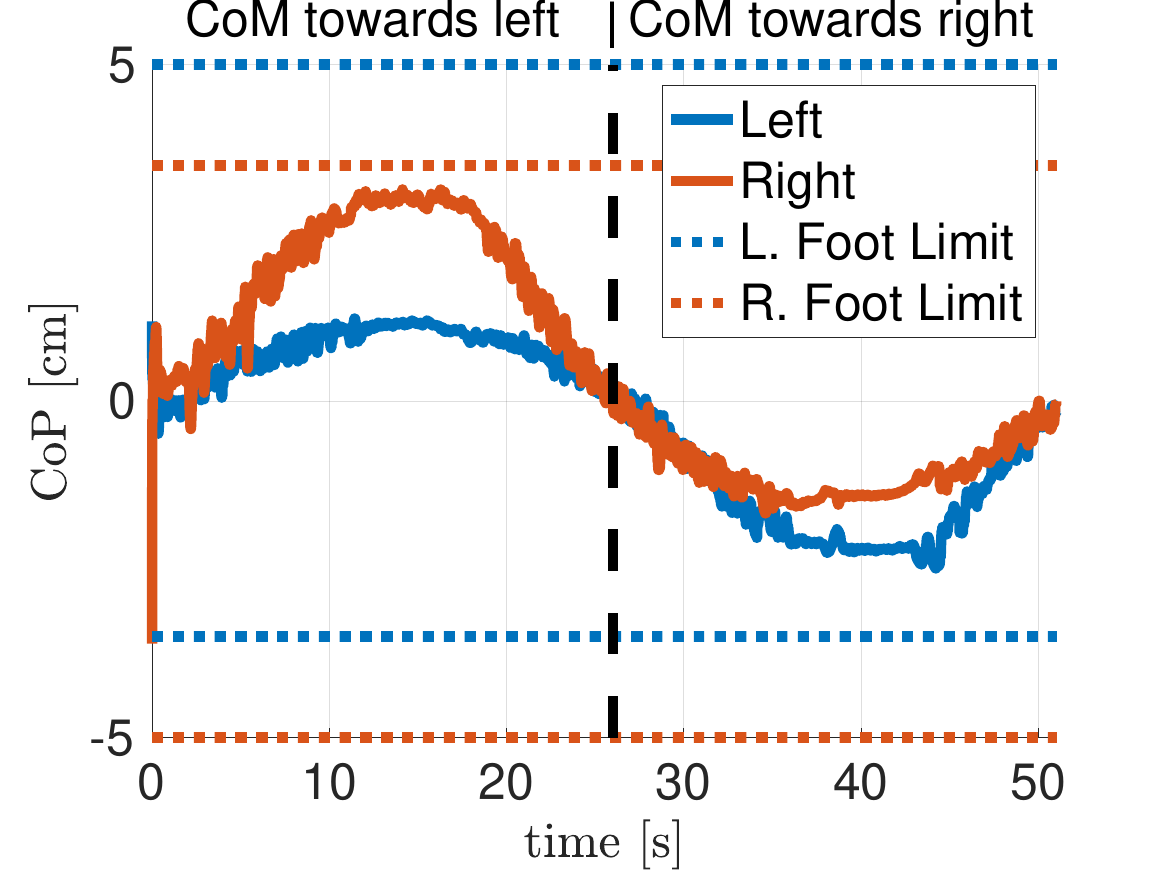}
    \caption{Center of pressure for the left and right foot when the minimisation of the joint  torques norm criterion is applied.}
    \label{fig:cop_min_torques}
\end{figure}

In this section, we validate the analysis performed on the four-bar linkage model on the full iCub humanoid robot.
We apply the same principles stated in criterion 1) and 2) in Section \ref{sec:choice_of_wrenches} to achieve balancing.

Our test platform is the iCub robot, a state-of-the-art $53$ degrees of freedom humanoid robot \cite{Metta20101125}. 
For the purpose of the present experiment, we consider only the principal $23$ degrees of freedom, located in the legs, torso and upper arms.
These degrees of freedom are supplied of a low level torque control that runs at $1\mathrm{kHz}$ and is responsible of stabilising desired joint torques.
As a comparison with the simplified model, the hips are about $50 \mathrm{cm}$ high and the feet are kept, during the experiments, at about $14 \mathrm{cm}$ of distance between each other.
Contact forces with the ground are estimated by  proper estimation algorithm that exploits six six-axes force torque sensors installed within iCub \cite[Sec. 3.2]{Nori2015}.

The controller is a dynamic controller composed of a strict hierarchy of two control objectives. 
The first, and with highest priority, is responsible of tracking a desired robot momentum through the use of the contact wrenches. 
Note that the desired robot momentum contains feedback terms on the center of mass position and velocity.
The second one, instead, is responsible of stabilising the resulting zero dynamics.
Discussion about the stability of the zero dynamics is out of the scope of the present paper and we refer the interested reader to \cite{Nava16} for further details on the zero dynamics stability problem.

Assuming only two contacts located at the feet, the dynamics of the robot momentum is given by the following equation:
\begin{equation}
    \label{eq:robotMomentumFull}
    \dot{H} = X_L f_L + X_R f_R + m \bar{g}
\end{equation}
where $H \in \mathbb{R}^6$ is the robot linear and angular momentum,
\[
    X_L = \begin{bmatrix}
        1_3 & 0_{3\times 3} \\
        S(x_L - x_{\text{CoM}})     & 1_3
    \end{bmatrix},~X_R = \begin{bmatrix}
        1_3 & 0_{3\times 3} \\
        S(x_R - x_{\text{CoM}})     & 1_3
    \end{bmatrix}
\]
are transformation matrices, $x_{\text{CoM}}$ is the position of the center of mass and $x_L, x_R$ are the positions of the left and right foot respectively. Finally $m\bar{g}$ is the wrench due to gravity.

Solutions to Eq. \eqref{eq:robotMomentumFull} yield the desired $f_L$ and $f_R$ to be applied by the robot, which are in turn generated by the actuation torques so as to satisfy the first control objective, i.e. tracking a desired robot momentum.
To obtain the torques to be applied, we can invert Eq. \eqref{eq:contact_forces}. 
If we denote with $f^* = [f_L^{*\top}, f_R^{*\top}]^\top$ a possible solution to Eq. \eqref{eq:robotMomentumFull}, we obtain the following expression for the control variable $\tau$:
\begin{equation}
    \tau = (J M^{-1} B)^\dagger [J M^{-1} (C\nu + G - J^\top f^*) - \dot{J}\nu],
\end{equation}
where $(J M^{-1} B)^\dagger$ denotes the Moore-Penrose pseudoinverse of the matrix $J M^{-1} B$.
The choice of $f^*$, and as a consequence the torques applied by the robot, depends on the criterion chosen to obtain a solution to Eq. \eqref{eq:robotMomentumFull}.
Similarly as we did for the four-bar linkage system we apply criterion {1)} and {2)} (see Sec. \ref{sec:choice_of_wrenches}) on the real robot and we compare the obtained results.

%
In the experiments, we move the robot along the frontal plane\footnote{We define with \emph{frontal plane} any vertical plane that divides the robot into anterior and posterior portions, similarly as in humans anatomy.} with very low velocity so as to simulate the quasi-static scenario presented in the analysis in Section \ref{sec:four-bar}.
We begin from the initial, symmetrical configuration, moving the robot towards the configuration where the CoM lies on the left or right foot frame origin. 
In particular, we impose a sinusoidal reference for the center of mass $y$-component, being the $y$-axis the axis along the robot frontal plane, and we fix the reference amplitude and frequency to be equal for the control laws associated with both criteria.
When we apply the criterion 2), we manage to track a full period of the reference sine, for the criterion 1) we have to perform two different experiments, i.e. one moving the CoM towards the left foot and the other towards the right one.

We first tested the minimum wrenches norm solution, i.e. criterion 1).
As we showed in the analysis in the previous section we expect that the vertical components of the forces do not change so much from the half weight of the robot.
This is confirmed also on the real robot. As we can see in Figure~\ref{fig:robot_minforces_forces} the forces start symmetrical. When the CoM starts moving towards the left foot we notice that both the forces change of only a limited quantity. 
The same behaviour was verified when we moved the CoM towards the right foot.
We then tested criterion 2), i.e. we exploited the redundancy of the forces to optimise for the joint torques.
Figure~\ref{fig:robot_mintorques_forces} shows the forces during the experiment for a full period of the CoM sinusoidal reference, highlighting the dependence of the forces upon the center of mass position.
%

We then analysed the center of pressure behaviour during the experiments.
Figure \ref{fig:cop_min_wrenches} shows the left and right foot center of pressure when we applied the first criterion. As mentioned previously we had to perform two different experiments, one where the CoM moves towards the left foot, in Fig. \ref{fig:cop_min_wrenches_leftcom}, and one where the CoM moves towards the right foot, in Fig. \ref{fig:cop_min_wrenches_rightcom}.
As the analysis on the four-bar linkage showed the two centers of pressure move almost in parallel regardless of which foot supports most of the weight.

When we apply the control law obtain by satisfying criterion 2), i.e. minimisation of the internal torques norm, the left and right foot centers of pressure behave differently, with the CoP of the foot supporting most of the weight moving less that the other one, as Figure \ref{fig:cop_min_torques} clearly highlights.
 

\begin{figure*}[t]
  \centering
\subfloat[Four-bar linkage model]{\includegraphics[width=.95\columnwidth]{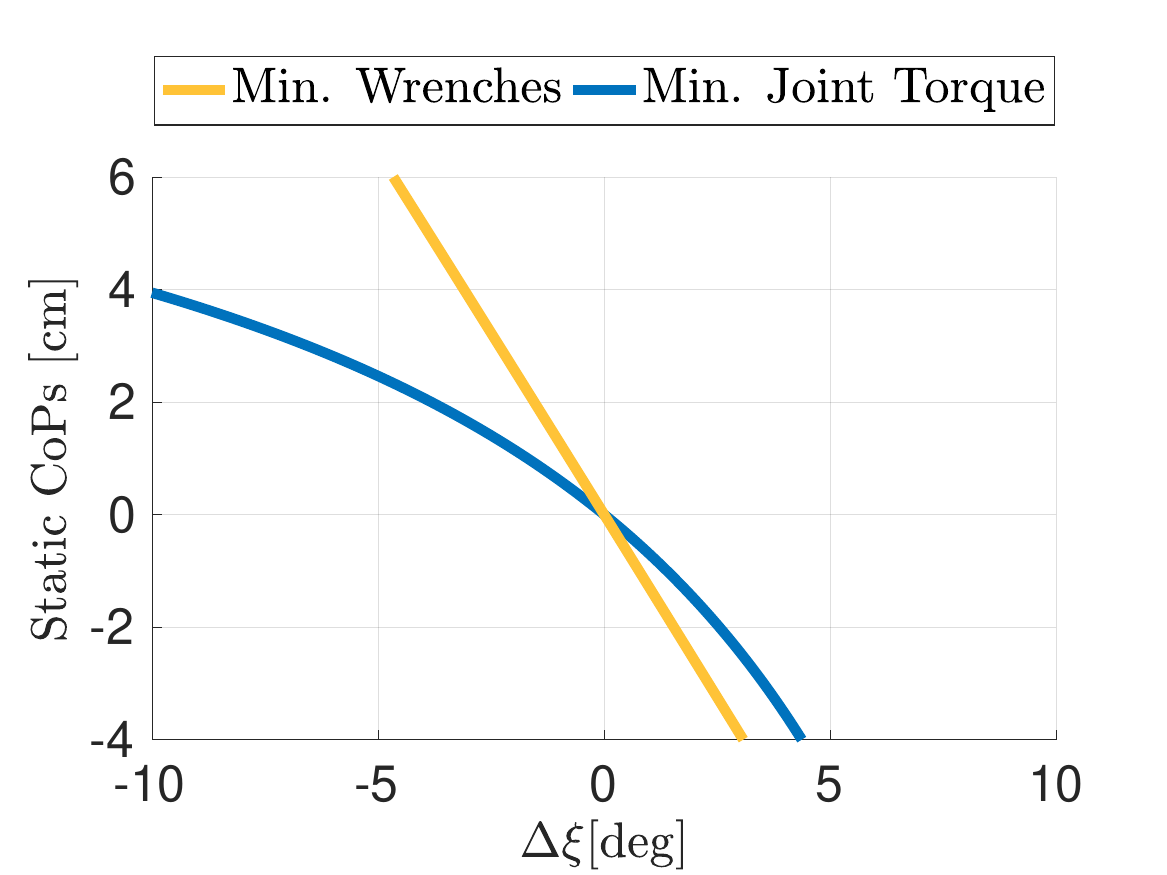}\label{fig:cop_comparison_leftcom_fourbar}} \subfloat[Robot]{\includegraphics[width=.95\columnwidth]{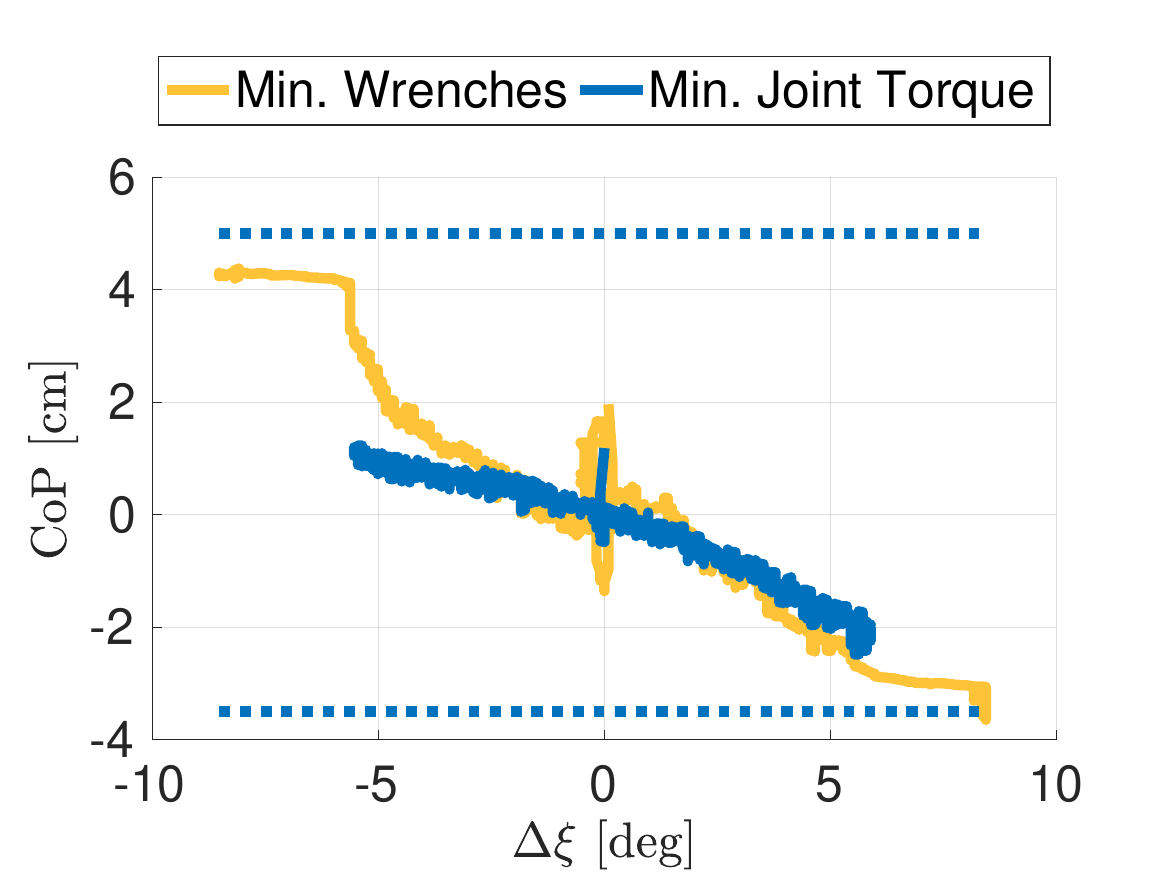}\label{fig:cop_comparison_leftcom_robot}}
    \caption{Comparison of the center of pressure of the left foot for the two criteria on the simplified model \protect\subref{fig:cop_comparison_leftcom_fourbar} and on the real robot \protect\subref{fig:cop_comparison_leftcom_robot}. The dotted blue lines in Figure~\protect\subref{fig:cop_comparison_leftcom_robot} denote the convex hull of the left foot.}
    \label{fig:cop_comparison}
\end{figure*}

\begin{Remark}
    As last step in our experimental validation, we compare how the two different criteria behave with respect to the minimal coordinate variables.
    Given the kind of movements performed by the robot, we assume as minimum coordinate variable $\xi$ the left hip roll joint, which corresponds to the joint $q_1$ in the four-bar linkage model shown in Figure \ref{fig:simple_model}.
    Having performed this choice, we can now properly compare the CoPs obtained with the two force criteria. 
    Fig. \ref{fig:cop_comparison} shows the aforementioned comparison considering only the left foot, where Fig \ref{fig:cop_comparison_leftcom_fourbar} shows the analytical results on the four-bar linkage model and Fig. \ref{fig:cop_comparison_leftcom_robot} shows the results collected on the iCub humanoid robot.
    It is worth noting the similarity between the two results, thus proving the validity of the reduced model we adopted for the theoretical analysis.
\end{Remark}

%


%

\section{Conclusions}
\label{sec:conclusions}

In this paper we introduced the concept of \emph{static center of pressure sensitivity} as an additional tool to analyse and evaluate the stability of a contact and to benchmark the performance of balancing controllers.
We first analysed the newly introduced metric on a planar four-bar linkage, which approximates the lower body of a biped robot when balancing.
Subsequently, we verified the analysis on the full iCub humanoid robot, obtaining similar results, thus proving the power of the approximate model.
The effect on the stability of the contacts of two different choices of contact forces are then shown and compared.

The analysis we performed, both analytically and numerically, shows how different criteria imply very different properties on the contact wrenches, and in turn, on the resulting center of pressures.
In particular the wrenches obtained by minimising the internal joint torques produces a lower sensitivity w.r.t. the minimum wrench norm solution.

We want to stress the fact that we do not advocate for the use of the proposed \emph{static center of pressure sensitivity} as a substitute to other criteria such as the center of pressure or the foot rotation indicator. On the contrary, the newly introduced metric can be used as an additional tool to discriminate between apparently similar balancing controllers.

During the analysis we noticed an additional, interesting property of the solution which minimises the joint torques.
Differently from the other solution, the vertical components of the contact force on one foot tends to zero when the center of mass of the robot moves towards the opposite foot.
While a proper theoretical analysis should be put into effort, this solution can help minimise discontinuities due to a change of the contact set.

\section*{Appendix}

\subsection{Proof of Lemma \ref{lemma:min_wrench}}\label{app:lemma1}
Consider the equilibrium condition of the centroidal dynamics \cite{Orin2013} of the mechanism, i.e. the balance of the external wrenches acting on the mechanism when written w.r.t. a frame centered in the center of mass and with the orientation of the inertial frame.
\begin{align}
    \label{eq:momentum_balance}
    0 & = X_L \prescript{L}{}f_L + X_R \prescript{R}{}f_R - m g e_2 = X f - mge_2
\end{align}
where $mge_2$ is the wrench due to gravity and $X := \begin{bmatrix}
X_L & X_R
\end{bmatrix}$,
\begin{align*}
    X_L &= \begin{bmatrix}
    1_2 & 0_{2\times1} \\ (S\lambda)^\top & 1
    \end{bmatrix}, \\
    X_R &= \begin{bmatrix}
    1_2 & 0_{2\times1} \\ (S\rho)^\top & 1
    \end{bmatrix},
\end{align*}
$\lambda = \begin{bmatrix}
\lambda_1 & \lambda_2
\end{bmatrix}^\top \in \mathbb{R}^2$ and $\rho = \begin{bmatrix}
\rho_1 & \rho_2
\end{bmatrix}^\top \in \mathbb{R}^2$ the vectors connecting the center of mass to the left and right foot frames respectively, see Figure \ref{fig:fourBarLinkageStanding}. 

Since $X \in \mathbb{R}^{3 \times 6}$ by construction is always full row rank, \eqref{eq:momentum_balance} admits infinite solutions, i.e.
\begin{equation}
\label{eq:4bar_minforces}
f = X^\dagger
mge_2 + N_f f_0
\end{equation}
where $X^\dagger \in \mathbb{R}^{6\times3}$ is the Moore-Penrose pseudoinverse of $X$, $N_f \in \mathbb{R}^{6\times6}$ is the projector onto the nullspace of $X$ and $f_0 \in \mathbb{R}^6$ is a free variable.
The unique minimum norm solution corresponds to the choice $f_0 = 0$.
The actual expression of the contact wrenches are thus the ones given in Eq. \eqref{eq:minforces_wrenches}.

The center of pressure in the planar case possesses only one coordinate, which is given by the following expressions
\begin{equation*}
{}^L\text{SCoP}_L = \frac{e_3^\top f}{e_2^\top f}, \quad {}^R\text{SCoP}_R = \frac{e_6^\top f}{e_5^\top f}
\end{equation*}
which evaluated with the expression of the wrenches in Eq. \eqref{eq:minforces_wrenches} yield
\begin{align*}
{}^L\text{SCoP}_L &= \frac{12 l \cos(\xi)}{5 d^2 + 6 l d \cos(\xi) + 20},\\ 
 {}^R\text{SCoP}_R &= \frac{12 l \cos(\xi)}{5 d^2 - 6 l d \cos(\xi) + 20}.
\end{align*}

To evaluate the static center of pressure sensitivity, we can differentiate the above expression w.r.t $\xi$, that is
\begin{align*}
\eta_L &= \frac{\partial ~\text{SCoP}_R}{\partial \xi} = -60 l \sin(\xi) \frac{d^2 + 4}{(5 d^2 + 6 l d \cos(\xi) + 20)^2},\\
\eta_R &= \frac{\partial ~\text{SCoP}_R}{\partial \xi} = -60 l \sin(\xi) \frac{d^2 + 4}{(5 d^2 - 6 l d \cos(\xi) + 20)^2}. \qed
\end{align*}

\subsection{Proof of Lemma \ref{lemma:min_torques}}
\label{app:lemma3}

Consider the balance of the external wrenches acting on an articulated rigid body in the general $3$D case:
\begin{align}
        \label{eq:momentum_balance3D}
      X_L \prescript{L}{}f_L + X_R \prescript{R}{}f_R - m g e_3 = 0
\end{align}
where $X_L,X_R \in \mathbb{R}^{6\times6}$ are used to express the wrenches respectively in $L$ and $R$ at the frame located at center of mass with the orientation of the inertial frame.
Impose the following expression for the contact wrenches:
\begin{equation}
    \label{eq:wrenches_feet}
    \begin{aligned}
        \prescript{L}{} f_L &= X_L^{-1} \frac{mg}{2}e_3 + X_L^{-1} \Delta \\
        \prescript{R}{} f_R &= X_R^{-1} \frac{mg}{2}e_3 - X_R^{-1} \Delta
    \end{aligned}  
\end{equation}
where $\Delta \in \mathbb{R}^6$ is a free variable.
We can notice that with the above choice, \eqref{eq:momentum_balance3D} is always satisfied independently of the choice of $\Delta$.
In what follows, the redundancy $\Delta$ is used to minimize the joint torques.
To simplify the calculations we perform a state transformation as described in \cite{Traversaro2016}.
In particular, the dynamics in the new state variables is decoupled in the acceleration, i.e. the mass matrix is block diagonal, with the two blocks being the terms relative to the base and the joints.
We refer the reader to \cite{Traversaro2016} for additional details, and we report here only the relevant transformations.
To avoid confusion, all the dynamic quantities in the new state variables are denoted with the overline. 

The new base frame is chosen to correspond to a frame with the same orientation of the inertial frame $\mathcal{I}$ and the origin instantaneously located at the center of mass.
We denote with ${}^c X_\mathcal{B}$
 the corresponding $6$D transformation.
Of particular interest is the form of the Jacobian of the left (right) foot frame $L$ ($R$) in the new state variables:
\begin{align*}
    \label{eq:jacobian_new}
    \xoverline{J}_{{L}} &= \begin{bmatrix}
        \xoverline{J}_{{L,b}} & \xoverline{J}_{{L,j}}
    \end{bmatrix} = \begin{bmatrix}
        X_{{L}}^\top & {J}_{{L,j}} - \frac{1}{m}X_{{L}}^\top \Lambda {M}_{bj}
    \end{bmatrix}, \\
    \xoverline{J}_{{R}} &= \begin{bmatrix}
        \xoverline{J}_{{R,b}} & \xoverline{J}_{{R,j}}
    \end{bmatrix} = \begin{bmatrix}
        X_{{R}}^\top & {J}_{{R,j}} - \frac{1}{m}X_{{R}}^\top \Lambda {M}_{bj}
    \end{bmatrix}, \\
    \Lambda &:=  1_6  - P ( 1_6 + {}^c X_B^{-1} P)^{-1}{}^cX^{-1}_B \notag.
\end{align*}
$P$ has the following expression:
\begin{equation*}
    P := \begin{bmatrix}
        0_{3\times3} & 0_{3\times3} \\
        S(p_{\text{CoM}} - p_\mathcal{B}) & \frac{I}{m} - 1_3 
    \end{bmatrix}
\end{equation*}
where $I$ is the $3$D inertia of the robot expressed w.r.t. the orientation of $\mathcal{I}$ and the center of mass as origin and $p_{\text{CoM}} - p_\mathcal{B}$ is the distance between the center of mass and the base frame before the state transformation.

The relation between the joint torques and the contact wrenches is given by Eq.~\eqref{eq:contactforcesEq}. So, by inverting this relation, we evaluate the effects of the contact wrenches redundancy $\Delta$ on the joint torques at the equilibrium configuration, i.e.
%
\begin{equation}
    \label{eq:tau_general_system}
    \tau = (\xoverline{J}_j \xoverline{M}_j^{-1})^\dagger \xoverline{J} \xoverline{M}^{-1} (mge_3 - \xoverline{J}^\top f).
\end{equation}
We can now substitute the definition of $f$ as in \eqref{eq:wrenches_feet} into the obtained expression of $\tau = \tau(f)$ in \eqref{eq:tau_general_system}, i.e. 
\begin{IEEEeqnarray}{rCl}
    \tau &=& 
     (\xoverline{J}_j \xoverline{M}_j^{-1})^\dagger \xoverline{J} \xoverline{M}^{-1} [ (2 - \xoverline{J}_L^\top X_L^{-1} - \xoverline{J}_R^\top X_R^{-1}) \frac{mg}{2}e_3 \notag\\
    && - (\xoverline{J}_L^\top X_L^{-1} - \xoverline{J}_R^\top X_R^{-1}) \Delta]. \label{eq:torques_with_delta}
\end{IEEEeqnarray}
Now, it is easy to verify that
\begin{align*}
    \xoverline{J}_L^\top X_L^{-1} + \xoverline{J}_R^\top X_R^{-1} &= \begin{bmatrix}
        2~1_6 \\ 
        {J}_{L,j}^\top X_L^{-1} + {J}_{R,j}^\top X_R^{-1} - 2 \frac{{M}_{bj}^\top}{m}\Lambda^\top
    \end{bmatrix} \\
    \xoverline{J}_L^\top X_L^{-1} - \xoverline{J}_R^\top X_R^{-1} &= \begin{bmatrix}
        0_{6 \times 6} \\
        {J}_{L,j}^\top X_L^{-1} - {J}_{R,j}^\top X_R^{-1}
    \end{bmatrix}.
\end{align*}

Observe also that 
\begin{equation*}
    P^\top e_3 = \begin{bmatrix}
        0_{3\times 3} & -S(r) \\
        0_{3\times 3} & \frac{I}{m}-1_3
    \end{bmatrix}e_3 = 0_6 \Rightarrow \Lambda^\top e_3 = 1_6.
\end{equation*}
Then \eqref{eq:torques_with_delta} becomes 
\begin{IEEEeqnarray}{rCl}
    \label{eq:torques_expression_full}
   \tau &=& - \left(\xoverline{J}_j \xoverline{M}_j^{-1} \right)^\dagger \xoverline{J}_j \xoverline{M}_j^{-1} \left[ \left({J}_{L,j}^\top X_L^{-1} + {J}_{R,j}^\top X_R^{-1} \right. \right. \notag\\
    &&\left. \left. - 2 \frac{{M}_{bj}^\top}{m} \right)\frac{mg}{2}e_3 + \left({J}_{L,j}^\top X_L^{-1} - {J}_{R,j}^\top X_R^{-1}\right)\Delta \right].
\end{IEEEeqnarray}

If we consider the four-bar linkage planar model, we can notice that the number of rows of $J_j$ are greater than the DoFs of the system and thus the product of the first two terms simplifies into the identity matrix. 
As a consequence, the vector $\Delta$ that minimizes the joint torques is given by:
\begin{IEEEeqnarray}{rCl}
    \Delta &=& -\frac{mg}{2} \left({J}_{L,j}^\top X_L^{-1}- {J}_{R,j}^\top X_R^{-1}\right)^\dagger \notag \\
    && \left({J}_{L,j}^\top X_L^{-1} + {J}_{R,j}^\top X_R^{-1}- 2 \frac{{M}_{bj}^\top}{m}\right)e_2 \label{eq:delta_torque_min}
\end{IEEEeqnarray}
which, by using the actual expression for the Jacobians, boils down to
\begin{equation}
    \label{eq:delta_simplified_model}
    \Delta = \frac{mg}{2} \begin{bmatrix}
        \frac{m_l + m_b}{m} \frac{l \cos^2(\xi)}{d \sin(\xi)}\\
        \frac{m_l + m_b}{m}  \frac{l\cos(\xi)}{d}\\
        \frac{d}{2}
    \end{bmatrix}.
\end{equation}
The complete expression of the contact wrenches can be obtained by substituting the expression of $\Delta$ in Eq. \eqref{eq:delta_simplified_model} into the following equation
    \begin{subequations}
         \begin{equation*}
            {}^L f_L = X_L^{-1} \frac{mg}{2} e_2 + X_L^{-1} \Delta,
         \end{equation*}
         \begin{equation*}
             {}^R f_R = X_R^{-1} \frac{mg}{2} e_2 - X_R^{-1} \Delta,
         \end{equation*}
     \end{subequations}
yielding the following final expression for the contact wrenches:
    \begin{subequations}
         \begin{align}
            {}^L f_L &= mg \left( \begin{bmatrix}
                0 \\ \frac{1}{2} \\ -\frac{\lambda_1}{2}
            \end{bmatrix} \right. \notag\\
            &+ \left. \begin{bmatrix}
                \frac{m_b + m_l}{m} \frac{l\cos(\xi)^2} {2 d \sin({\xi})} \\
                \frac{m_b + m_l}{m} \frac{l \cos({\xi})}{2d} \\
                \frac{m_b + m_l}{m} \frac{l \cos({\xi})}{2d} \left( \lambda_2 \frac{\cos(\xi)}{\sin({\xi})} - \lambda_1 \right) + \frac{d}{4}
            \end{bmatrix} \right),
         \end{align}
         \begin{align}
             {}^R f_R &= mg \left( \begin{bmatrix}
                 0 \\ \frac{1}{2} \\ -\frac{\rho_1}{2}
             \end{bmatrix} \right. \notag\\
             &- \left. \begin{bmatrix}
                 \frac{m_b + m_l}{m} \frac{l\cos(\xi)^2} {2 d \sin({\xi})} \\
                 \frac{m_b + m_l}{m} \frac{l \cos({\xi})}{2d} \\
                 \frac{m_b + m_l}{m} \frac{l \cos({\xi})}{2d} \left( \rho_2 \frac{\cos(\xi)}{\sin({\xi})} - \rho_1 \right) + \frac{d}{4}
             \end{bmatrix} \right).
         \end{align}
     \end{subequations}
Note that by assuming also Hypothesis \ref{hyp:four_bar}-\emph{iii)} one obtains the expressions in Eq. \eqref{eq:mintorques_wrenches}.

We can now compute the center of pressures $\text{SCoP}_L$ and $\text{SCoP}_R$.
Taking the forces expression in \eqref{eq:wrenches_feet}, the static CoPs are thus:
\begin{equation}
    \label{eq:cop_expression}
    \begin{aligned}
        \text{SCoP}_L & = \frac{-\frac{1}{2}\rho_1 + \rho_2 \Delta_1 - \rho_1 \Delta_2 + \Delta_3}{\frac{1}{2} + \Delta_2},\\
        \text{SCoP}_R & = \frac{-\frac{1}{2}\rho_1 - \rho_2 \Delta_1 + \rho_1 \Delta_2 - \Delta_3}{\frac{1}{2} - \Delta_2}.
    \end{aligned}    
\end{equation}

By plugging the expression of $\Delta$ in Eq. \eqref{eq:delta_simplified_model} we obtain the expression of the static center of pressure as a function of the free variable $\xi$, which is
\begin{equation}
    \label{eq:cop_expression_final}
    \begin{aligned}
        \text{SCoP}_L & = \frac{9ld\cos(\xi)}{20d + 6l\cos(\xi)}, \\
        \text{SCoP}_R & = \frac{9ld\cos(\xi)}{20d - 6l\cos(\xi)}.
    \end{aligned}    
\end{equation}

To obtain the expression of the sensitivity of the static center of pressure we can differentiate Eq. \eqref{eq:cop_expression_final} w.r.t $\xi$, finally obtaining
\begin{align*}
    \eta_L &= \frac{\partial ~\text{SCoP}_L}{\partial \xi} = - \frac{45d^2l\sin(\xi)}{(10d + 3l\cos(\xi))^2}\\
   \eta_R &= \frac{\partial ~\text{SCoP}_R}{\partial \xi} = - \frac{45d^2l\sin(\xi)}{(10d - 3l\cos(\xi))^2}.\qed
\end{align*}

\subsection{Proof of Lemma \ref{lemma:min_tang}}\label{app:lemma2}

    Consider Eq. \eqref{eq:4bar_minforces}. 
    We want to use $f_0$ to minimize the tangential component of the contact wrenches, i.e. we want find the solution to
    \begin{equation*}
        \begin{bmatrix}
            e_1^\top \\
            e_4^\top 
        \end{bmatrix} f = 0 \iff \begin{bmatrix}
            e_1^\top \\
            e_4^\top 
        \end{bmatrix} N_f f_0 + \begin{bmatrix}
            e_1^\top \\
            e_4^\top 
        \end{bmatrix} X^\dagger mge_2 = 0.
    \end{equation*}
    Because of the expression of the nullspace projector and of the pseudoinverse, the above equation reduces to
    \begin{equation*}
        \frac{1}{2}
        \begin{bmatrix}
            1 & 0 & 0 & -1 & 0 & 0 \\
            -1 & 0 & 0 &  1 & 0 & 0
        \end{bmatrix} f_0 = 0
    \end{equation*}
    which has the minimum norm solution in $f_0 = 0$.
\qed

\bibliographystyle{IEEEtran}
\bibliography{IEEEabrv,references}

\end{document}